\RequirePackage[svgnames]{xcolor}

\documentclass[11pt,letterpaper]{mystyle}
\usepackage[all]{hypcap}
\usepackage[svgnames]{xcolor}
\usepackage[comma,authoryear,compress]{natbib}
\hypersetup{
    colorlinks = true,
    citecolor = {YaleBlue},
}

\usepackage{algorithm}
\usepackage{algorithmicx}
\usepackage{algpseudocode}
\usepackage{microtype}
\usepackage{graphicx}
\expandafter\def\csname ver@subfig.sty\endcsname{}
\usepackage{booktabs} %
\usepackage{float}
\usepackage{bigstrut}

\usepackage{amsmath}
\usepackage{amssymb}
\usepackage{mathtools}
\usepackage{amsthm}
\usepackage{mathrsfs}
\usepackage{nicefrac}
\usepackage{dsfont}
\usepackage{subcaption}
\usepackage{graphicx,subfig}
\usepackage{cleveref}
\usepackage{bxcoloremoji}
\usepackage{listings}
\usepackage{tcolorbox}
\tcbuselibrary{listings}

\newtcblisting{promptbox}[2][]{
    listing only,
    listing options={
        basicstyle=\ttfamily\footnotesize,
        breaklines=true,
    },
    colback=gray!10,
    colframe=YaleBlue!60, 
    title=#2,
    fonttitle=\bfseries,
    #1
}
\usepackage{enumitem}
\setlist[itemize]{leftmargin=*}
\setlist[enumerate]{leftmargin=*}
\usepackage[normalem]{ulem}

\usepackage{float}
\usepackage{multirow}
\usepackage{afterpage}

\usepackage[utf8]{inputenc} %
\usepackage[T1]{fontenc}    %
\usepackage{url}            %
\usepackage{booktabs}       %
\usepackage{amsfonts}       %
\usepackage{nicefrac}       %
\usepackage{microtype}      %
\usepackage{graphicx}
\usepackage{subcaption} 
\usepackage{amssymb}
\usepackage{fdsymbol}
\usepackage{wrapfig}
\usepackage{lipsum}
\usepackage{stackengine}
\usepackage[font=small,labelfont=bf]{caption}
\usepackage{color}
\usepackage{adjustbox}

\usepackage{rotating}
\usepackage{array}
\usepackage{arydshln}
\usepackage{makecell}
\usepackage[flushleft]{threeparttable}
\usepackage{listings}  
\providecolor{HeaderGray}{HTML}{EFEFEF}
\providecolor{GroupGreen}{RGB}{220,240,220}
\providecolor{GroupBlue}{RGB}{91,157,218}
\providecolor{GroupOrange}{RGB}{255,235,200}
\providecolor{mySoftBlue}{RGB}{205,230,255}
\providecolor{myExtremelyLightBlue}{RGB}{235,245,255}
\providecolor{cvprblue}{rgb}{0.21,0.49,0.8}

\providecommand{\cmark}{\textcolor{green!60!black}{\ding{51}}}
\providecommand{\xmark}{\textcolor{red!75!black}{\ding{55}}}
\providecommand{\pmark}{\textcolor{GroupBlue}{\ding{51}\rotatebox[origin=c]{-6.2}{\kern-0.7em\ding{55}}}}

\newcommand{\x}{\mathbf{x}}

\definecolor{blanchedalmond}{rgb}{1.0, 0.92, 0.8}
\definecolor{carmine}{rgb}{0.59, 0.0, 0.09}
\definecolor{lightblue}{rgb}{0.22,0.45,0.70}%

\renewcommand{\mathbf}{\boldsymbol}

\makeatletter
\def\Ddots{\mathinner{\mkern1mu\raise\p@
\vbox{\kern7\p@\hbox{.}}\mkern2mu
\raise4\p@\hbox{.}\mkern2mu\raise7\p@\hbox{.}\mkern1mu}}
\makeatother

\definecolor{amaranth}{rgb}{0.9, 0.17, 0.31}
\definecolor{antiquebrass}{rgb}{0.8, 0.58, 0.46}
\definecolor{antiquefuchsia}{rgb}{0.57, 0.36, 0.51}
\definecolor{chromeyellow}{rgb}{0.31, 0.47, 0.26}

\newtcolorbox{AIbox}[2][]{aibox,title=#2,#1}
\definecolor{lightblue}{rgb}{0.22,0.45,0.70}%
\definecolor{Gray}{gray}{0.95}
\definecolor{Cornsilk}{rgb}{1.0, 0.97, 0.86}

\usepackage{amsmath}

\usepackage{pdfrender}

\definecolor{Gray}{gray}{0.95}
\newcolumntype{a}{>{\columncolor{Gray}}c}

\usepackage{skak}
\usepackage{bbm}
\usepackage{lipsum}
\usepackage{listings}
\usepackage{caption}
\usepackage{fancyhdr}

\definecolor{pink}{HTML}{fc6c85}

\definecolor{customblue}{HTML}{286dc0}

\definecolor{customred}{HTML}{CC0000}

\definecolor{customyellow}{HTML}{ffd55a}

\definecolor{customgrey}{HTML}{978d85}

\newtcolorbox{blueBox}[1][]{
  colback=YaleBlue!12!white,  
  colframe=YaleBlue!85,   
  boxrule=1pt,
  arc=3pt,
  floatplacement=floating,
  title=\centering #1
}

\newtcolorbox{yellowBox}[1][]{
  colback=customyellow!10!white,
  colframe=customyellow,
  floatplacement=floating,
  title=\centering #1
}

\newtcolorbox{promptBox}[1][]{
  colback=YaleBlue!12!white,  
  colframe=YaleBlue!85,      
  boxrule=0.8pt,
  arc=3pt,
  floatplacement=floating,
  title=\centering #1
}

\newtcolorbox{wronganswer}[1][]{
    enhanced,
    breakable,
    colframe=customred!65,
    colback=customred!6!white,
    sharp corners,
    boxsep=0pt,
    left=5pt,
    right=5pt,
    top=6pt,
    bottom=6pt,
    boxrule=0pt,
    leftrule=4pt,
    #1
}

\newtcolorbox{correctanswer}[1][]{
    enhanced,
    breakable,
    colframe=OliveGreen,
    colback=OliveGreen!10!white,
    sharp corners,
    boxsep=0pt,
    left=5pt,
    right=5pt,
    top=6pt,
    bottom=6pt,
    boxrule=0pt,
    leftrule=4pt,
    #1
}

\usepackage{fontawesome5}

\title{\textbf{WorldMemArena}: Evaluating Multimodal Agent Memory Through Action–World Interaction}
\runningtitle{\textbf{WorldMemArena}: Evaluating Multimodal Agent Memory Through Action–World Interaction}

\settitleleftlogo[2.2cm]{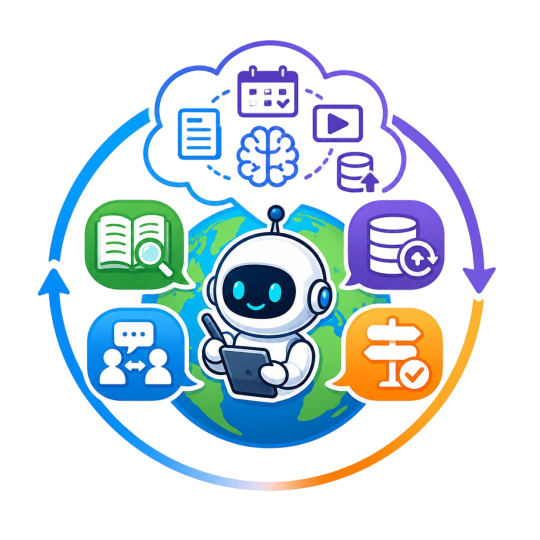} 
\settitleleftlogogap{-2.2mm}                   
\settitleleftlogooffset{3mm}{-3mm}     
\settitleboxverticalpadding{5mm}{5mm}    
\settitlespacing{5pt}{11pt}{15pt}             
\settitlebottomrightlogos{}                    

\newcommand{\affilicon}[1]{%
  \raisebox{-0.18em}{\includegraphics[height=1.1em]{#1}}%
}

\makeatletter
\def\@author{%
  {\Authfont
  Chengzhi Liu*\textsuperscript{1},
  Yuzhe Yang*\textsuperscript{1},
  Sophia Xiao Pu\textsuperscript{1},
  Yepeng Liu\textsuperscript{1},
  Lin Long\textsuperscript{5},
  Yichen Guo\textsuperscript{1},
  Nuo Chen\textsuperscript{6},\\
  Zhaotian Weng\textsuperscript{1},
  Elena Kochkina\textsuperscript{2},
  Simerjot Kaur\textsuperscript{2},
  Charese Smiley\textsuperscript{2},
  Xiaomo Liu\textsuperscript{2},
  James Zou\textsuperscript{4},\\
  Sheng Liu\textsuperscript{4},
  Yuheng Bu\textsuperscript{1},
  Songyou Peng\textsuperscript{3},
  Xin Eric Wang\textsuperscript{1}\par}
  \vspace{0.6em}
  {\Affilfont
\textsuperscript{1}\affilicon{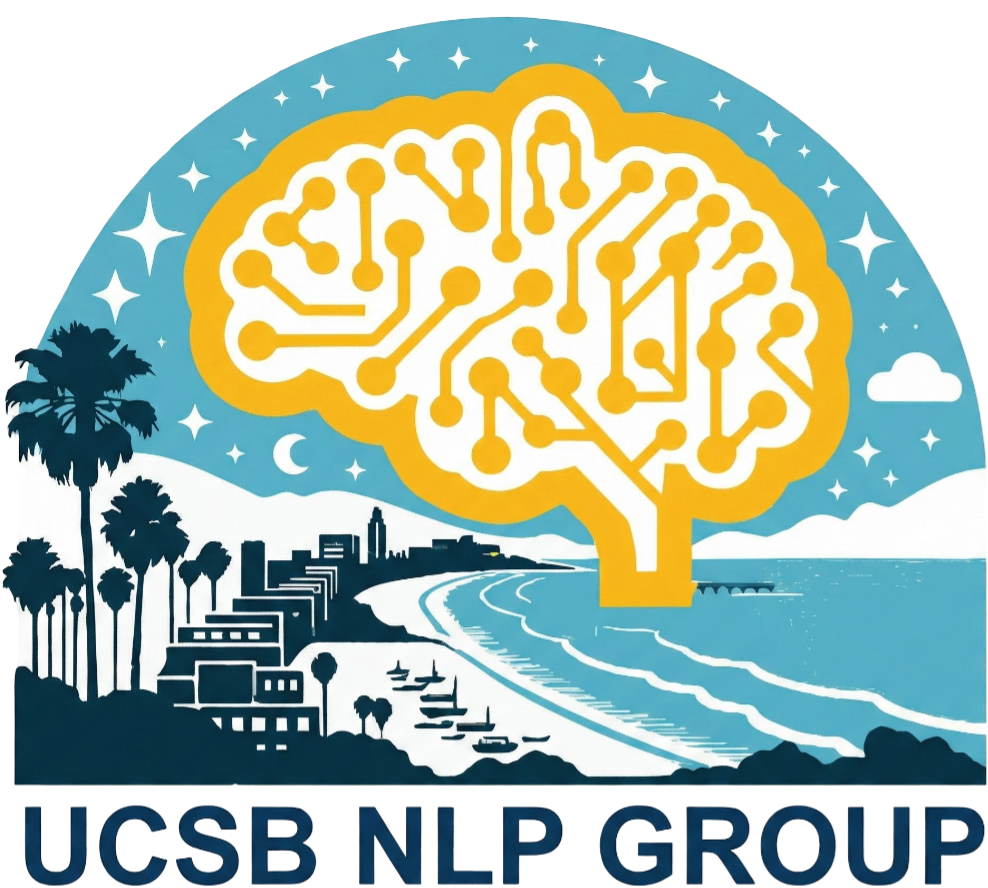}~University of California, Santa Barbara\quad
\textsuperscript{2}\affilicon{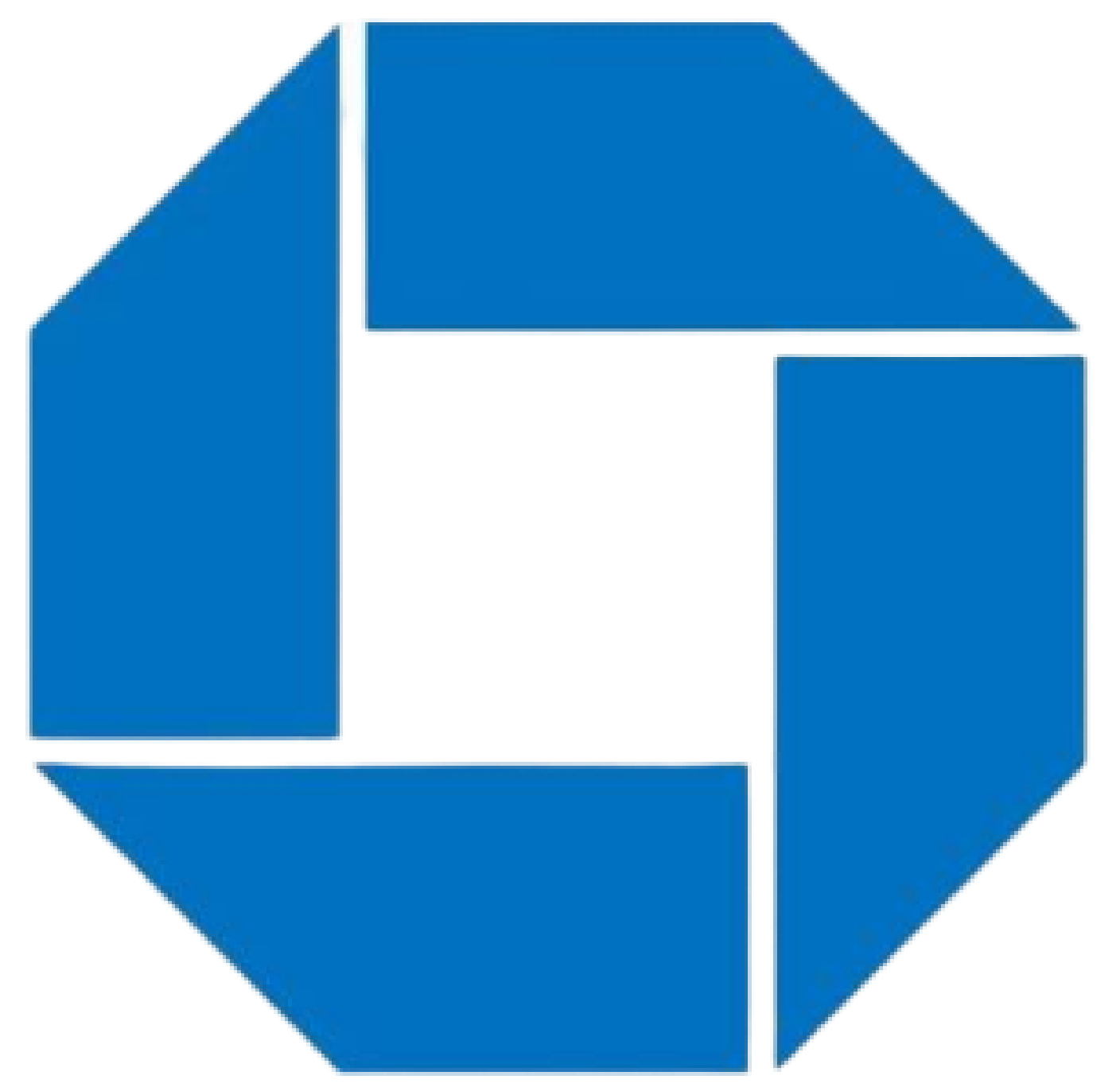}~J.P.~Morgan Chase\quad \textsuperscript{3}\affilicon{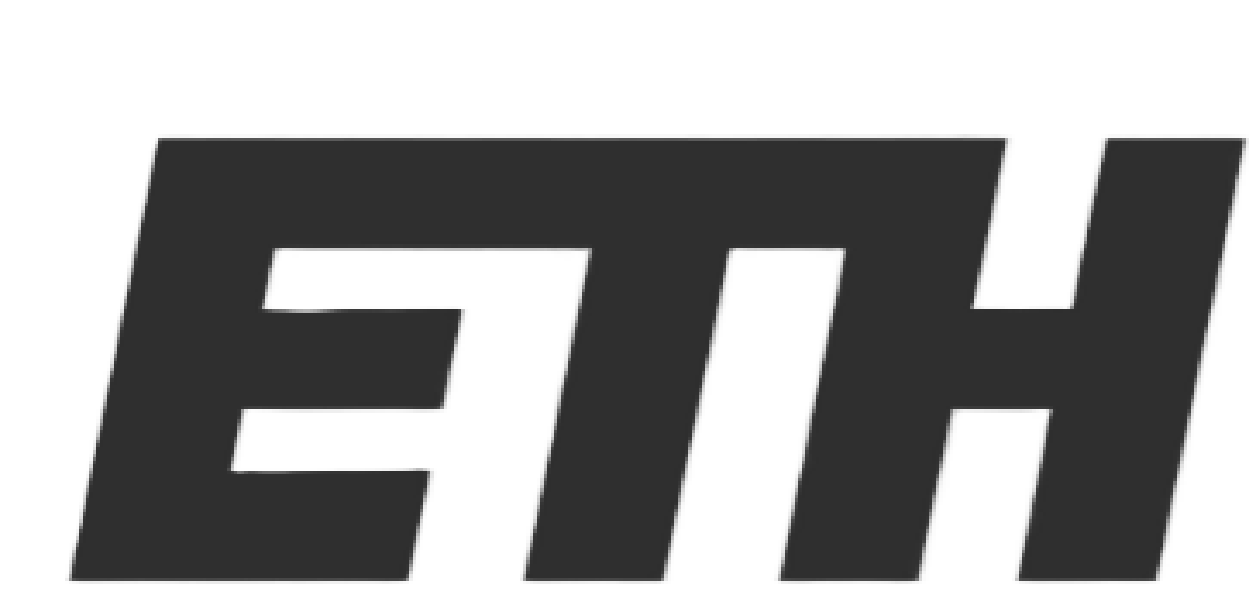}~ETH Zurich\\
\textsuperscript{4}\affilicon{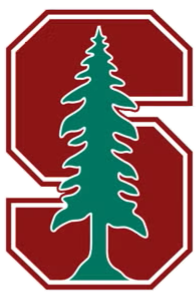}~Stanford University\quad
\textsuperscript{5}\affilicon{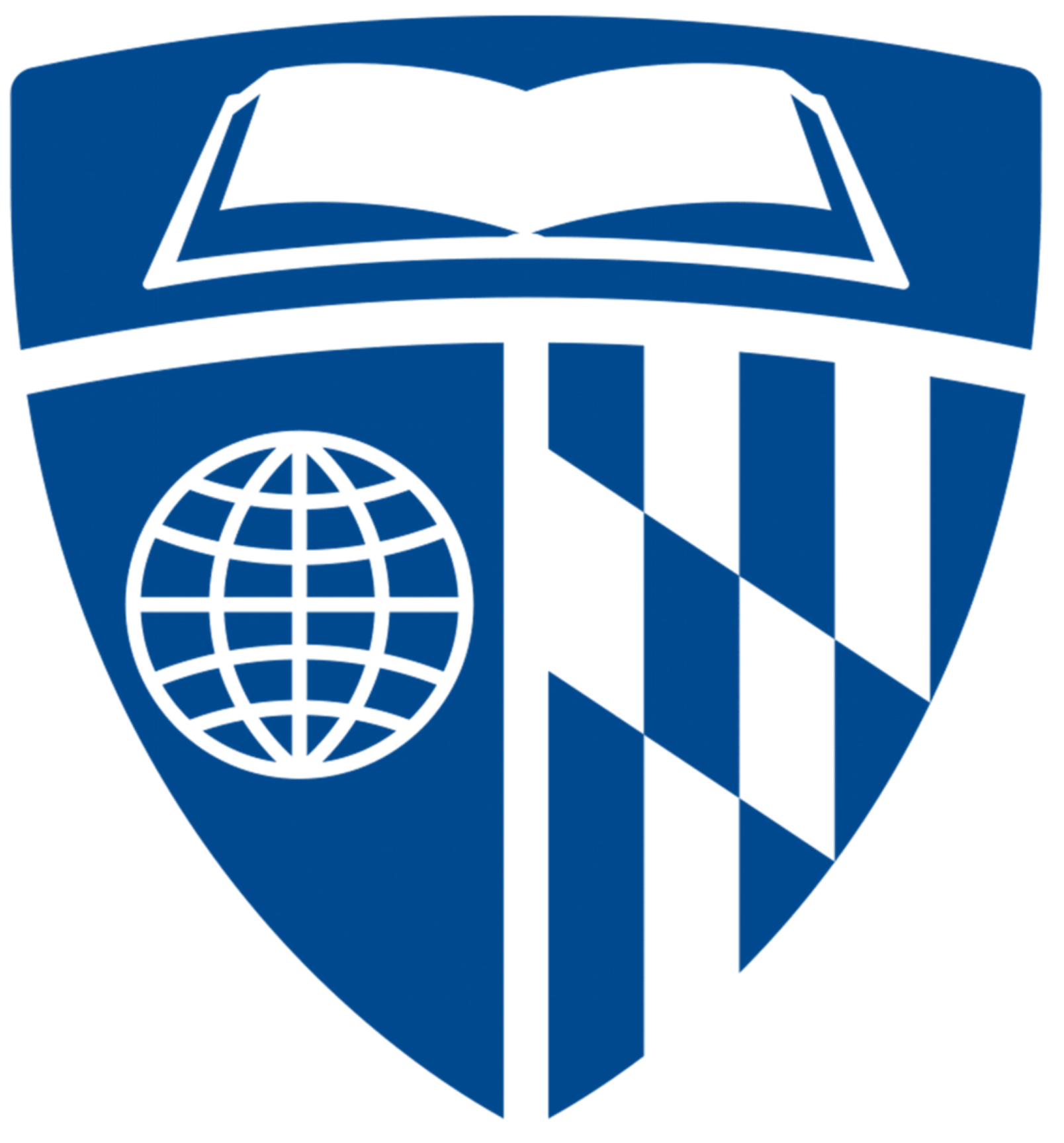}~Johns Hopkins University\quad
\textsuperscript{6}\affilicon{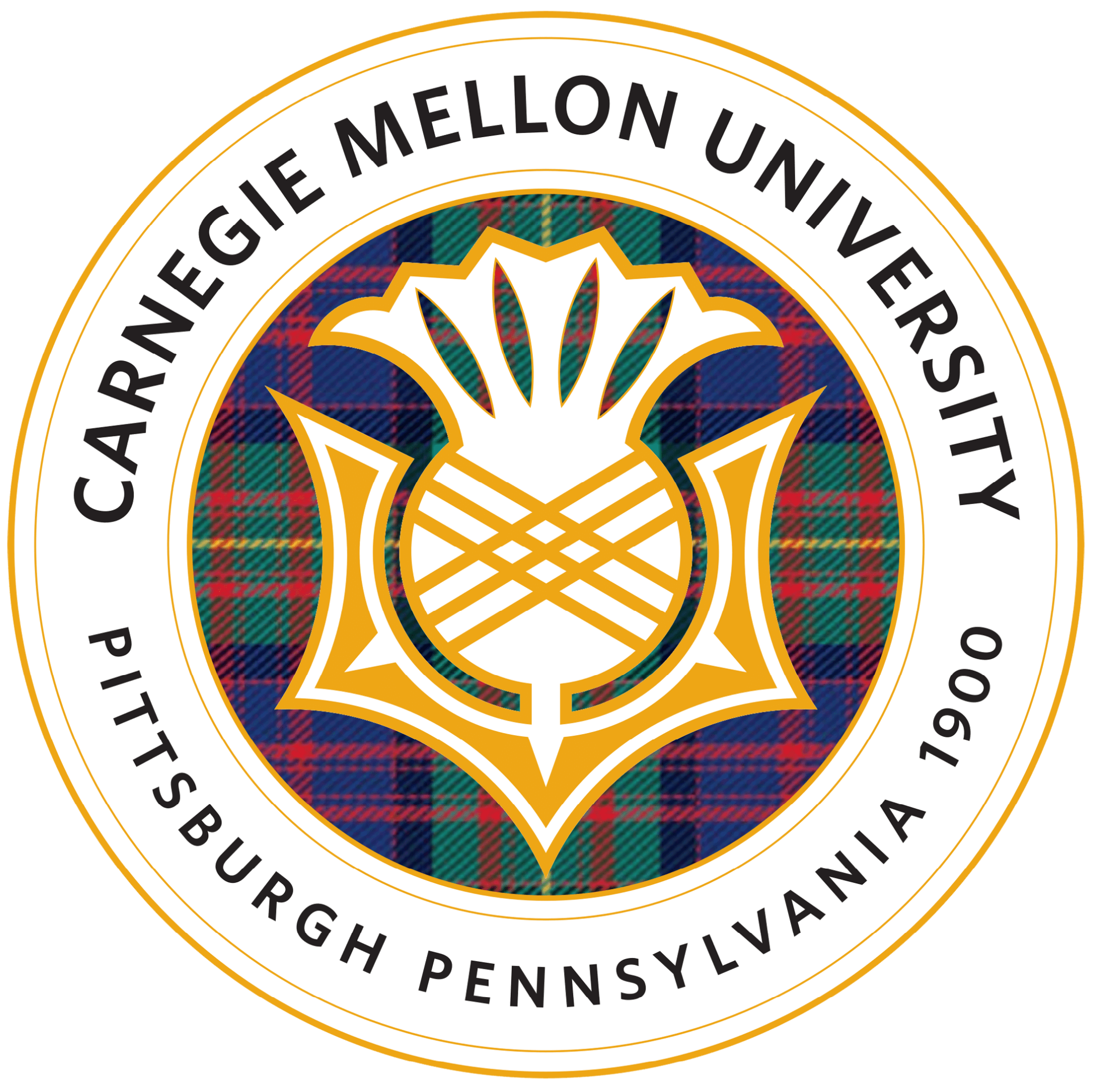}~Carnegie Mellon University\par}
}
\makeatother

\begin{document} 

\begin{abstract}
Multimodal large language models are increasingly deployed as long-horizon agents, where memory must do more than recall: it must track an evolving world, revise what has gone stale, and surface the right evidence at decision time. Existing benchmarks measure recall over static dialogue, collapse memory into a single end-of-task accuracy, and reduce visual observations to captions, leaving us unable to localize failures to writing, maintenance, retrieval, or use. The rise of agent harnesses that author their own memory sharpens this gap, since we have no principled way to compare hand-designed pipelines with self-managing alternatives. To close these gaps, we formulate multimodal agent memory as an \textit{Action--World Interaction Loop} with an observable four-stage lifecycle, and instantiate it in \textbf{WorldMemArena}: 461 multi-session multimodal tasks spanning \textit{Lifelong Evolution} (evolving personal and task states) and \textit{Agentic Execution} (memory from real observations, actions, and feedback), annotated with gold memory points, updates, distractors, and evidence chains for stage-level diagnosis. This enables the first head-to-head comparison of long-context, manually designed (RAG and external memory systems), and harness-based memory agents. Results show that:
\textit{\textbf{(1)}} better memory writing and storage do not guarantee better performance; \textit{\textbf{(2)}} multimodal memory still struggles to fully use visual evidence; \textit{\textbf{(3)}} systems are unstable across domains and degrade on realistic agentic trajectories; and \textit{\textbf{(4)}} harness memory is more flexible but remains costly and less reliable.
\vspace{4mm}

\parbox{\linewidth}{
\raisebox{-0.05em}{\faEnvelope}\hspace{0.3em}\textbf{Correspondence: }\texttt{\{chengzhi,yuzheyang,ericxwang\}@ucsb.edu}\\
\raisebox{-0.1em}{\faGlobe}\hspace{0.3em}%
\href{https://worldmemarena-mem.github.io/}{\textbf{Project Page}}%
\hspace{0.6em}%
\raisebox{-0.2em}{\includegraphics[height=1em]{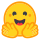}}\hspace{0.2em}%
\href{https://huggingface.co/datasets/LCZZZZ/WorldMemArena}{\textbf{Dataset}}%
\hspace{0.6em}%
\href{https://github.com/UCSB-AI/WorldMemArena}{
  {\color[RGB]{63,120,185}%
    \raisebox{-0.1em}{\faGithub}%
    \hspace{0.3em}\textbf{WorldMemArena}%
  }%
}
}
\end{abstract}

\maketitle

\begin{figure}[H]
\centering
\includegraphics[width=0.82\linewidth]{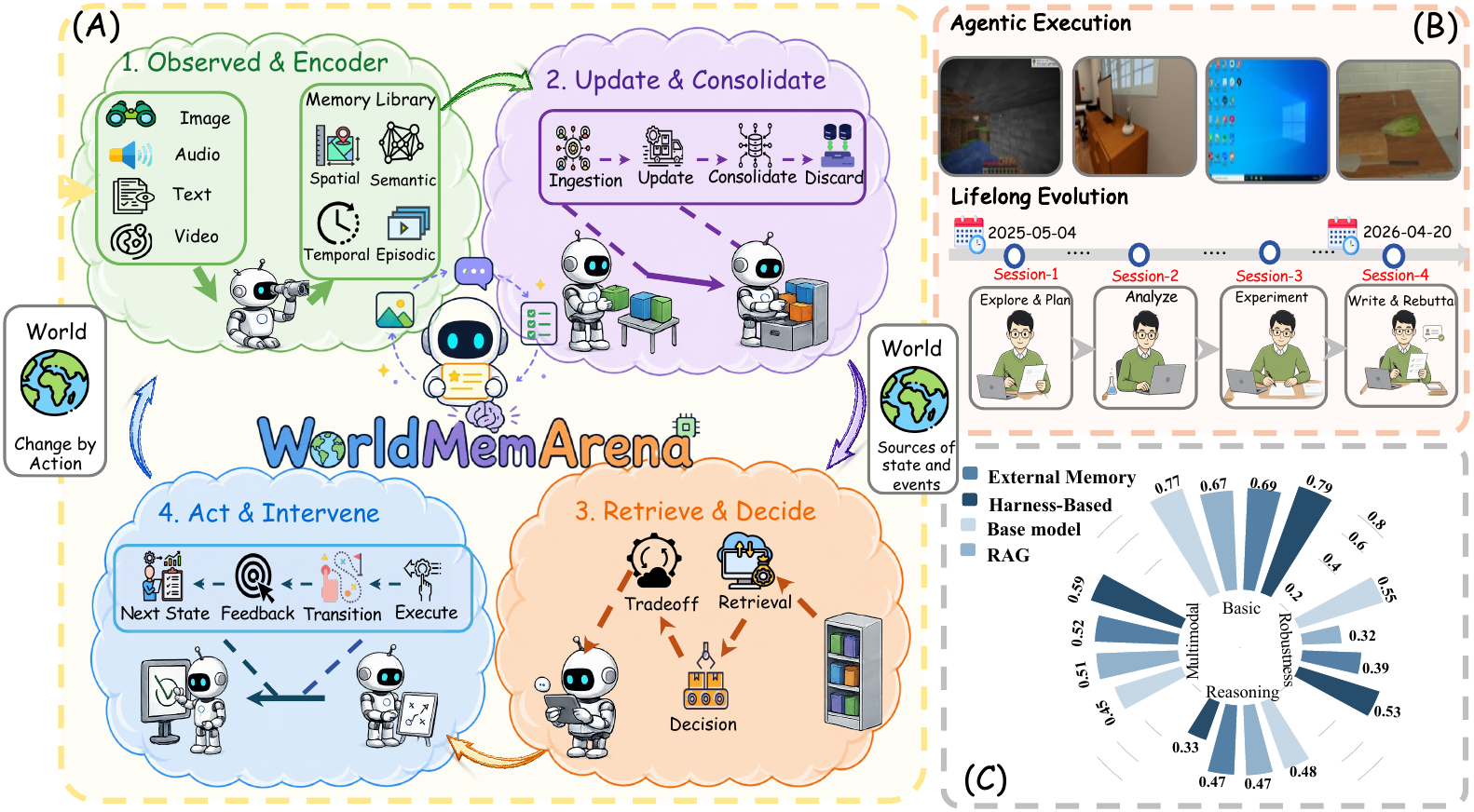}
\vspace{-0.6em}
\caption{\fontsize{9.3pt}{10pt}\selectfont (a) WorldMemArena formulates multimodal agent memory as an \textit{Action-World Interaction Loop}, where agents write observations, update evolving memory, retrieve evidence for decisions, and act in the world with feedback. (b) It spans two regimes, \textit{Agentic Execution} and \textit{Lifelong Evolution}, covering real agent trajectories and evolving personal and task states across sessions.  (c) Evaluation covers different memory paradigms across basic, robustness, reasoning, and multimodal capabilities.}
\label{fig:teaser}
\vspace{-1em}
\end{figure}

\section{Introduction}
\label{sec:intro}

Multimodal large language models~\cite{gpt54,qwen35_2026, claude_opus46_2026} are turning from question answering systems into agents that act in dynamic environments over long horizons~\cite{steinberger2025openclaw, claudecode2026}.  In this setting, memory is no longer simply a cache of past text, but a mechanism for tracking task state, learning from actions, and supporting decisions  through real-world interaction. A capable long horizon agent should not only recall the past, but also write useful information, revise outdated memories, and retrieve the right evidence for future decisions. How well current memory systems can fulfill this role remains insufficiently evaluated.

\begin{wrapfigure}{r}{0.56\textwidth}
  \centering
  \vspace{-1em}
  \includegraphics[width=\linewidth]{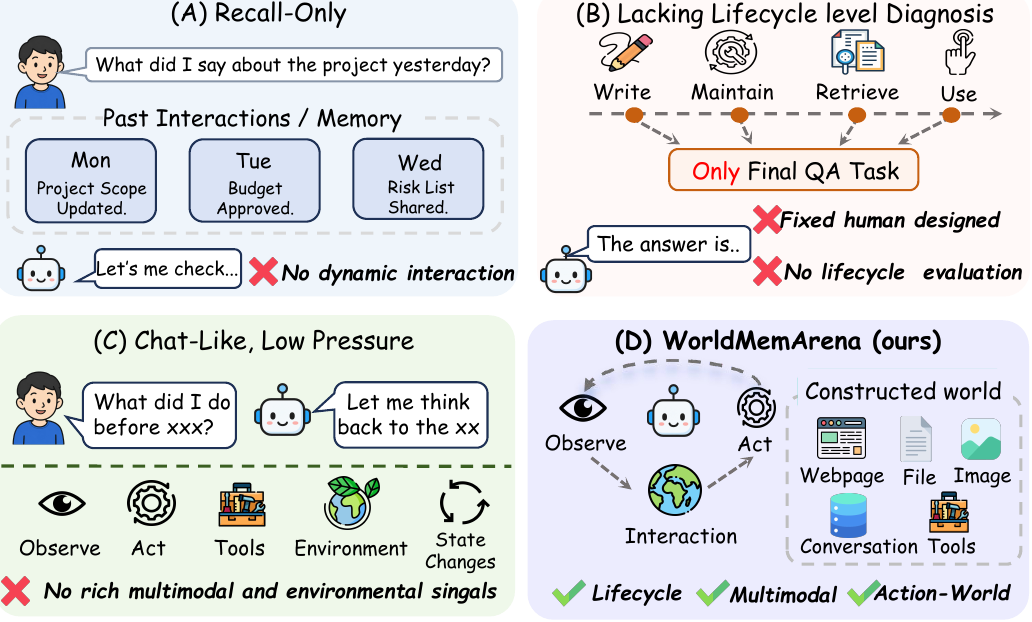}
  \caption{Overview of multimodal agent memory evaluation.
(a) Recall-only evaluation. (b) Missing lifecycle-level diagnosis.
(c) Low-pressure chat-like settings. (d) WorldMemArena evaluates multimodal memory through action-world interaction.}
  \label{fig:intro}
\end{wrapfigure}

Existing benchmarks fall short of this picture in three connected ways. \textbf{\textit{(i)}} They are often built around long dialogues or extended contexts~\cite{ jiayang2026amemgyminteractivememorybenchmarking}, testing what models can remember rather than how they use past experience to guide future actions.(Figure~\ref{fig:intro}(a)).  \textbf{\textit{(ii)}} As shown in Figure~\ref{fig:intro}(b), many evaluations~\cite{zhao2026amabenchevaluatinglonghorizonmemory, hu2026evaluatingmemoryllmagents, liu2025thinkingseeingassessingamplified} report only final question answering accuracy, without checking whether relevant evidence is written, updated, retrieved, and used at the right time, making it difficult to identify where memory failures occur. \textbf{\textit{(iii)}} Figure~\ref{fig:intro}(c) shows that existing benchmarks remain largely text-centric, often converting images into captions before evaluation, with limited real interaction and insufficient pressure on multimodal evidence use.

Beyond these evaluation limitations, current benchmarks also miss a deeper shift in how agent memory is built and used. Agent harness systems such as OpenClaw~\cite{steinberger2025openclaw} and Codex~\cite{codex2026} now let agents author and reorganize their own memory during interaction, blurring the line between the memory module and the policy that uses it. In the spirit of \textit{Sutton's Bitter Lesson}, this invites a question the field should be asking head-on:

\begin{tcolorbox}[colframe=black!50, colback=cvprblue!8, boxrule=1.5pt, arc=2mm, top=2pt, bottom=3pt, left=4pt, right=4pt,  boxsep=3pt]
\raisebox{-0.2\baselineskip}{\includegraphics[height=1\baselineskip]{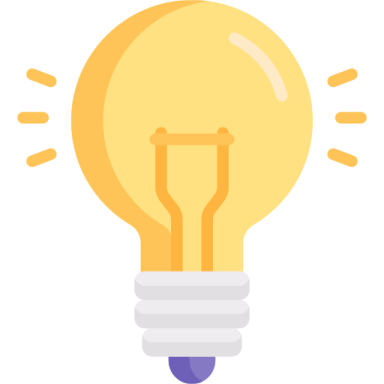}} \textit{ When agents can continuously act in realistic environments and manage their own experience, should agent memory be evaluated as a predefined write-update-retrieve pipeline, or as a capability formed through interaction and used to support future decisions over time?}
\end{tcolorbox}

Answering this question requires an evaluation that treats memory as a process rather than a static snapshot. As shown in Figure~\ref{fig:teaser}, we reframe multimodal agent memory as an \textit{\textbf{Action and World Interaction Loop}}. At each step, the agent observes a partially visible world, takes an action, receives feedback, and uses memory to guide future actions and retain useful evidence. Under this view, memory has an observable lifecycle that covers what is written, how it is maintained as the world changes, what evidence is retrieved, and how the retrieved evidence is used. As shown in Figure~\ref{fig:intro}(c), each stage can be evaluated using shared trajectory evidence, rather than inferred from a single accuracy score.

We instantiate this view in \textbf{WorldMemArena}, a multimodal multi-session benchmark of 461 long-horizon interaction tasks spanning two complementary regimes. \textit{Lifelong Evolution} focuses on personal and task states that evolve across sessions, requiring systems to continuously track, update, and reuse long-term memories. \textit{Agentic Execution} places memory in realistic agent trajectories, where systems must extract reusable evidence from observations, actions, and feedback rather than relying on pre-organized textual narratives. Each session is annotated with gold memory points, state updates, distractors, and answer supporting evidence chains. These annotations support diagnosis across memory writing, maintenance, retrieval, and use, while providing a shared evidence base for comparing different memory systems.

Under a unified setting, the evaluation covers long-context agents, manually designed memory systems, and memory agents built on execution harnesses. The results reveal four findings: 
\textit{\textbf{(1)}} Storing more correct memories does not guarantee better performance; the key is whether they can be used correctly at answer time. \textit{\textbf{(2)}} multimodal memory remains a major bottleneck, especially for complex visual reasoning tasks;
\textit{\textbf{(3)}} memory performance varies across domains and degrades on agentic execution tasks, where key information is distributed across actions, tool feedback, and state changes; and
\textit{\textbf{(4)}} manually designed memory systems are more structured but less adaptive, while harness based memory agents are more flexible but remain costly and less reliable. To sum up, our contributions are listed as follows:

\begin{itemize}[left=0.2cm]
    \item We formulate multimodal agent memory as an \textit{Action--World Interaction Loop} and define a four stage lifecycle of writing, maintenance, retrieval, and use.

    \item We introduce \textbf{WorldMemArena}, a multi-session multimodal benchmark covering \textit{Lifelong Evolution} and \textit{Agentic Execution}, with annotations for stage level memory diagnosis.

    \item We conduct a unified comparison of three representative agent memory paradigms, identifying their respective strengths, failure modes, and implications for future design.
\end{itemize}

\section{Related Works}
\label{sec:related}

\textbf{Memory Benchmarks and Evaluation.} Early memory benchmarks such as LoCoMo~\cite{maharana2024evaluating}, MemoryAgentBench~\cite{hu2025evaluating}, and Realme~\cite{bian2026realmembenchmarkingllmsrealworld} focus on long-dialogue settings, measuring whether models can retain and recall historical information. These benchmarks treat memory as static recall over text and do not capture how memory supports dynamic task execution. More recent agent-oriented benchmarks~\cite{he2026memoryarenabenchmarkingagentmemory, zhao2026amabenchevaluatinglonghorizonmemory, liu2024visualagentbenchlargemultimodalmodels} incorporate tool traces, environment feedback, and task dependencies, moving closer to realistic agent-environment interaction. However, evaluation still centers on final success rates or question answering accuracy, making it difficult to identify where and why memory fails. WorldMemArena differs by decomposing evaluation into writing, maintenance, retrieval, and use, making it possible to localize where memory failures originate.

\textbf{Multimodal Memory Mechanisms.} Recent multimodal memory systems~\cite{long2025seeinglisteningrememberingreasoning, liu2025memversemultimodalmemorylifelong, zhou2026videomemoryconsistentvideogeneration, fu2026latentmemcustomizinglatentmemory} have demonstrated strong capabilities in visual understanding and long-term information retention. Their evaluations, however, are largely confined to image and video comprehension tasks, with limited attention to how memory operates within agent interaction loops. Benchmarks that incorporate multimodal memory~\cite{bei2026memgallerybenchmarkingmultimodallongterm, lu2026mmamultimodalmemoryagent, yang2025embodiedbenchcomprehensivebenchmarkingmultimodal, wang2024mementoscomprehensivebenchmarkmultimodal, liu2026reasoningminddynamicmultimodal} extend evaluation to images, videos, and dialogues, but cover a narrow range of scenarios and apply limited evaluation pressure on evidence reuse. WorldMemArena broadens the scope to multi-session agent interaction, testing whether systems can preserve, update, and reuse multimodal evidence as tasks and environments evolve.

\section{Problem Formulation} 

\subsection{Memory as an Action-World Interaction Loop}
\label{sec:problem}
We define each instance as a long horizon agent-world interaction process. Given an initial task context $x$, the agent does not directly observe the full world state. At step $t$, the world has a latent state $z_t$, from which the agent receives an observation $o_t$. The agent then selects an action $a_t$ based on the observation and its current memory state $m_t$. After the action is executed, the environment updates its state and returns feedback $f_t$:
\vspace{-0.3em}
\[
o_t=\Omega(z_t), \qquad a_t=\pi(o_t,m_t), \qquad (z_{t+1},f_t)=\mathcal{E}(z_t,a_t).
\]
Here, $\Omega$ maps the latent world state to observable inputs, $\pi$ denotes the agent policy, and $\mathcal{E}$ represents the environment response, including both state transition and feedback generation. Observations may include  language, visual inputs or logs, while actions may include responses, tool calls, or execution.

Based on the above process, we denote the full trajectory as
$\tau=(x;\eta_1,\ldots,\eta_T)$, where each event
$\eta_t=(o_t,a_t,f_t)$ records the observation, action, and feedback at step
$t$. To evaluate long-horizon memory, we further segment the trajectory into
sessions, i.e.,
$\tau=\tau^{(1)}\circ\tau^{(2)}\circ\cdots\circ\tau^{(S)}$.
Within each session, the agent only observes local context, while the world state persists and evolves across sessions. This creates a natural point: later decisions may depend on evidence that is no longer directly visible, and we focus on whether the agent can recover and use such evidence through memory.

\subsection{Memory Lifecycle as a Diagnostic Framework}
\label{sec:problem-lifecycle}
The Action World Interaction Loop in \S\ref{sec:problem-lifecycle} is architecture agnostic. It does not assume where memory is stored or how it is represented. This allows us to evaluate different memory systems through four observable phases of writing, maintenance, retrieval, and use. These phases capture the shared lifecycle of preserving and reusing information across sessions.

\textbf{$\spadesuit$ Observe to Write.} This phase evaluates  \textit{whether the system can identify future useful evidence
from the current session.} Given the previous memory state $m_{s-1}$ and the
current session trajectory $\tau^{(s)}$, the system produces a memory delta
$\Delta_s=\textsc{Write}(m_{s-1},\tau^{(s)})$. The objective is selective
retention, keeping information that may support future responses or actions
rather than storing the full trajectory.

\textbf{$\spadesuit$ Update and Consolidate.} This phase evaluates \textit{how newly written information is integrated into
existing memory.} The system updates its state as
$m_s=\textsc{Maintain}(m_{s-1},\Delta_s)$.  Since long-horizon interaction is
not purely additive, memory must support revision and consolidation as user
preferences, task states, and environmental evidence evolve.

\textbf{$\spadesuit$ Retrieve for Decision.} This phase evaluates \textit{whether the system can access the right evidence when a
future query or decision need arises.} For a query $q$, retrieval returns
$R_{s,q}=\textsc{Retrieve}(m_s,q)$. The goal extends beyond semantic similarity to decision relevance, requiring the retrieved context to contain evidence needed for the current answer or action.

\textbf{$\spadesuit$ Use and Act.} This phase evaluates \textit{whether retrieved memory is faithfully used in the final
response or action.} Given $q$ and retrieved evidence $R_{s,q}$, the system
outputs $\hat{y}_{s,q}=\textsc{Answer}(q,R_{s,q})$. Failures may still arise
when the system ignores relevant evidence, relies on outdated memory, or fails
to translate prior experience into appropriate action.

\begin{table*}[t]
\centering
\small
\setlength{\aboverulesep}{0pt}
\setlength{\belowrulesep}{0pt}
\setlength{\tabcolsep}{3pt}
\renewcommand{\arraystretch}{1.15}
\caption{Comparison of \textbf{WorldMemArena} with representative memory benchmarks.
\texorpdfstring{\cmark}{full} = satisfies,
\texorpdfstring{\pmark}{partial} = partial support,
\texorpdfstring{\xmark}{absent} = does not satisfy.
\textit{MM} denotes multimodal support; \textit{Dim.}\ denotes evaluation dimensions;
\textit{\#QA} the number of QA pairs; \textit{Img.}\ the number of images;
\textit{Session} the number of multi-turn sessions; \textit{Steps} the number of interaction steps;
\textit{Mode} the interaction paradigm. The \textit{Lifecycle} group covers the
\textit{Write}, \textit{Update} (\textit{Upd.}), \textit{Retrieve} (\textit{Ret.}), and \textit{Use} stages of memory.}
\label{tab:bench-cmp-new}
\resizebox{\textwidth}{!}{%
\begin{tabular}{@{}lcccccccc cccc@{}}
\toprule
\rowcolor{HeaderGray}
& & & & & & & & &
\multicolumn{4}{c@{}}{\textbf{Lifecycle}} \\
\rowcolor{HeaderGray}
\multirow{-2}{*}{\textbf{Benchmark}}
& \multirow{-2}{*}{\textbf{MM}}
& \multirow{-2}{*}{\textbf{Dim.}}
& \multirow{-2}{*}{\textbf{Eval.}}
& \multirow{-2}{*}{\textbf{\#QA}}
& \multirow{-2}{*}{\textbf{Img.}}
& \multirow{-2}{*}{\textbf{Session}}
& \multirow{-2}{*}{\textbf{Steps}}
& \multirow{-2}{*}{\textbf{Mode}}
& \textbf{Write} & \textbf{Upd.} & \textbf{Ret.} & \textbf{Use} \\
\midrule
LoCoMo~\cite{maharana2024evaluating}
& \pmark & $5$ & Static & $1{,}986$ & $910$ & $272$ & $5{,}882$ & Dialogue
& \xmark & \xmark & \cmark & \cmark \\
LongMemEval~\cite{wu2024longmemeval}
& \xmark & $5$ & Static & $500$ & -- & $23{,}867$ & $246{,}750$ & Long-context
& \xmark & \cmark & \cmark & \cmark \\
MemoryAgentBench~\cite{hu2025evaluating}
& \xmark & $4$ & Static & $3{,}671$ & -- & $146$ & $6{,}484$ & Long-context
& \xmark & \cmark & \cmark & \cmark \\
MMRC~\cite{xue2025mmrc}
& \cmark & $6$ & Static & $2{,}105$ & $1{,}193$ & $457$ & $11{,}784$ & Dialogue
& \pmark & \cmark & \cmark & \cmark \\
HaluMem~\cite{chen2026halumemevaluatinghallucinationsmemory}
& \xmark & $3$ & Static & $3{,}467$ & -- & $1{,}387$ & $60{,}146$ & Dialogue
& \cmark & \cmark & \xmark & \cmark \\
RealMem~\cite{bian2026realmembenchmarkingllmsrealworld}
& \xmark & $4$ & Static & $1{,}415$ & -- & $2{,}055$ & $14{,}028$ &  Dialogue
& \xmark & \pmark & \cmark & \cmark \\
Mem-Gallery~\cite{bei2026memgallerybenchmarkingmultimodallongterm}
& \cmark & $3$ & Static & $1{,}711$ & $1{,}003$ & $240$ & $7{,}924$ & Dialogue
& \pmark & \cmark & \cmark & \cmark \\
AMA-Bench~\cite{zhao2026amabenchevaluatinglonghorizonmemory}
& \xmark & $4$ & Interactive & $2{,}496$ & -- & $208$ & $15{,}244$ & Agent
& \pmark & \cmark & \pmark & \cmark \\
MEMORYARENA~\cite{he2026memoryarenabenchmarkingagentmemory}
& \xmark & $4$ & Interactive & $4{,}850$ & -- & $701$ & $4{,}850$ & Agent
& \xmark & \xmark & \xmark & \cmark \\
\midrule
\rowcolor{myExtremelyLightBlue}
\textbf{WorldMemArena}
& \cmark & $27$ & \textbf{Interactive} & $24{,}258$ & $15{,}595$ & $8{,}489$ & $59{,}858$ & \textbf{Dialog.+Agent}
& \cmark & \cmark & \cmark & \cmark \\
\bottomrule
\end{tabular}%
}
\end{table*}

\section{\textit{WorldMemArena}: Agent Memory in Action-World Interaction}
\label{sec:benchmark}

\textbf{Overview.} WorldMemArena consists of 461 multi-session multimodal interaction tasks across two regimes (\textit{Lifelong Evolution} and \textit{Agentic Execution}). Each task is a temporally ordered sequence of sessions, where the agent receives partial observations and must rely on memory to inform decisions in later sessions. To support fine-grained diagnosis, every session is annotated with three types of structured labels. \textit{Gold memory points} specify the information that should be retained after a session, representing ground-truth memory content. \textit{State updates} mark where previously stored information becomes outdated and must be revised, testing whether the memory system can maintain temporal consistency. \textit{Distractors} introduce plausible but irrelevant or superseded information, testing whether the system can distinguish currently valid evidence from noise. In addition, each question is paired with \textit{evidence points}, the subset of gold memory points that are necessary to answer it correctly. These annotations together enable evaluation at each stage of the memory lifecycle.

\subsection{Memory Regimes}
\label{sec:regimes}

\textbf{Agentic Execution.} Each instance is derived from a real or realistic agent trajectory containing observations, actions, and environment feedback. Later steps depend on earlier outcomes, so the agent must convert past execution experience into reusable memory that informs future decisions.

\textbf{Lifelong Evolution.} Each instance is generated from a hidden world state that evolves across sessions. It covers two scenarios: (1) \textit{lifelong personal evolution}, where scattered interactions must be consolidated into coherent personal memory; and (2) \textit{long-horizon projects}, where task goals, intermediate results, and feedback shift across stages, requiring the agent to maintain up-to-date progress memory.

\textbf{Why both Regimes are Needed.} As the Action-World Interaction Loop requires the agent to both observe an evolving world and act within it, two demands on memory arise:
\textit{(1) Persistent state tracking} requires maintaining an accurate representation of an evolving world across sessions, which is evaluated by \textit{Lifelong Evolution} through controlled state evolution.
\textit{(2) Action grounded experience reuse} requires turning observations, action outcomes, and feedback into knowledge for later decisions, which is evaluated by \textit{Agentic Execution} through realistic execution trajectories.

\begin{figure}[t]
\centering
\includegraphics[width=\linewidth]{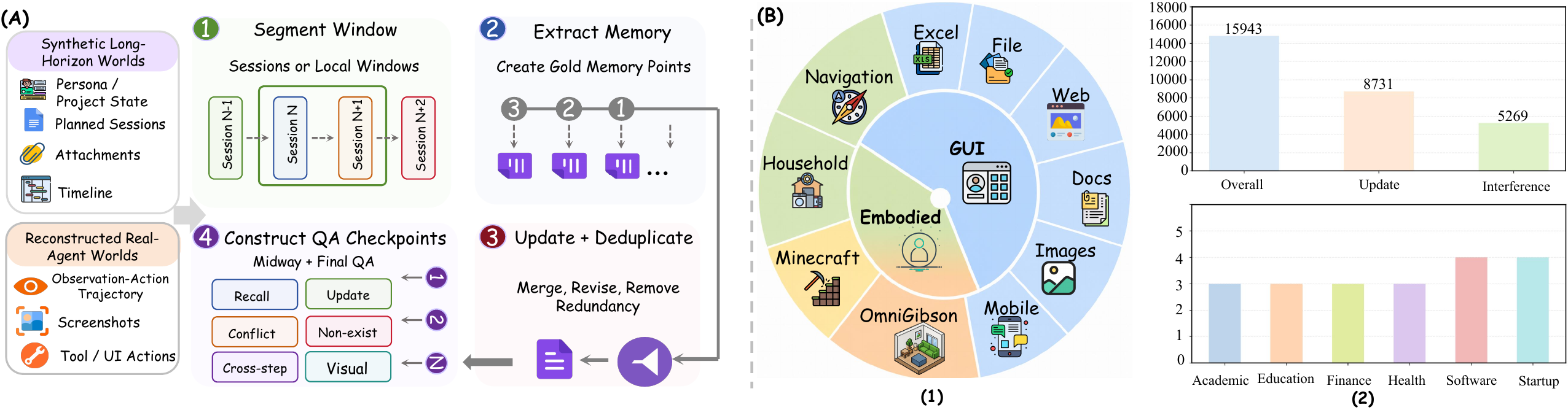}
\caption{Data construction pipeline and benchmark composition. (a) WorldMemArena constructs data around two task regimes, \textit{Lifelong Evolution} and \textit{Agentic Execution}, by segmenting sessions, extracting and updating gold memory points, removing redundancy, and constructing midway and final QA checkpoints. (b-1) The benchmark covers both GUI and embodied interaction settings. (b-2) The upper charts summarize gold memory points across the benchmark, including update memory points and interference memory points. The lower chart shows the task distribution across domains in \textit{Lifelong Evolution}.}
\label{fig:pipeline}

\end{figure}

\vspace{-0.1cm}
\subsection{Data Collection}
\vspace{-0.2cm}
As shown in Figure~\ref{fig:pipeline}(a), WorldMemArena is constructed through a unified automated memory construction pipeline with four steps.
\textit{{\textbf{(1)}}} Raw data is segmented into multi-session instances. For \textit{Lifelong Evolution}, a hidden world state is first defined and sessions are generated in temporal order, each revealing partial information about a persona or project. For \textit{Agentic Execution}, existing agent trajectories are split at subgoal boundaries, key feedback points, or state changes. \textit{\textbf{(2)}} For each session window, gold memory points are extracted, covering facts to retain, state updates to revise, and evidence required by future questions. \textit{\textbf{(3)}} Memory points are merged, revised, and deduplicated across sessions to remove redundancy and ensure temporal consistency.  \textit{\textbf{(4)}} Question-answer pairs are constructed from the refined gold memory points, covering 11 question types. Each instance is further reviewed by 2-3 human annotators to ensure quality.

\subsection{\textcolor{black}{Data Statics}}

\textbf{Dataset Scale and Coverage.} Table~\ref{tab:bench-cmp-new} compares WorldMemArena with existing benchmarks. Prior datasets typically focus on either long-form dialogue or agentic trajectories, whereas this benchmark covers both lifelong evolution and agentic execution. It contains 461 multi-session samples, with an average of 18.4 sessions and approximately 9.1K tokens per sample, making it substantially longer than existing multimodal memory benchmarks. It further provides 24{,}258 QA pairs and 15{,}595 images or screenshots, supporting broader question coverage and richer visual grounding. Most existing benchmarks do not evaluate the full memory lifecycle; the closest prior work, HaluMem, addresses memory storage and recall but remains limited to the textual modality.

\textbf{Domain and Annotations.} As shown in Figure~\ref{fig:pipeline}(b), \textit{Lifelong Evolution} covers 6 domain specific project types, with each session containing an average of 4 images and 15-20 dialogue turns. \textit{Agentic Execution} preserves real agent execution traces and their corresponding visual states, covering 6 GUI subcategories and 4 Embodied subcategories. Across both regimes, fine-grained lifecycle annotations are provided.
Each session contains an average of 10 key memory points, 3 update points,
and 2 interference points. Each sample further includes staged QA
checkpoints with an average of 5 evaluation positions. Each question is
paired with retrieval evidence, where most require 1-2 evidence items
and more complex questions require 5-6, covering both textual and visual information.

\subsection{Evaluation Protocol}
Following the four lifecycle stages defined in \S\ref{sec:problem-lifecycle}, we evaluate whether a memory system can correctly write, maintain, retrieve, and use memory across long horizon interactions. Detailed metric definitions and settings are provided in the Appendix~\ref{app:eval-retrieval}. 

\textit{\textbf{Stage 1.}} For each session, newly written memory items are matched against the gold memory points introduced in that session, with \textit{memory recall} used as the coverage metric. Each written item is further assessed by an LLM-as-a-Judge and classified as \textit{correct, hallucinated, or irrelevant}, distinguishing effective memory writing from noisy or unsupported storage.

\textit{\textbf{Stage 2.}} For gold memory points marked as updates, the system memory after the corresponding session is examined to determine whether the new information is preserved and the obsolete version is properly handled. \textit{An update is considered successful only when the revised memory is retained and the old version is removed or overwritten}. This criterion prevents simple accumulation of historical information from being misclassified as effective memory maintenance.

\textit{\textbf{Stage 3.}} For each checkpoint question, the retrieved memory items are matched against the annotated gold evidence. The evidence may be grounded in either textual or visual information, and all evidence types are evaluated under a unified coverage criterion. \textit{Recall} measures whether the required evidence is retrieved, while Normalized Discounted Cumulative Gain \textit{(NDCG)} measures whether relevant evidence is ranked near the top, thereby separating retrieval quality from final answer correctness.

\textit{\textbf{Stage 4.}} Checkpoint questions are grouped into four categories and twelve capability axes: \textit{Basic} covers factual recall; \textit{Robustness} covers dynamic update, memory boundary, and memory conflict; \textit{Reasoning} covers temporal reasoning, knowledge reasoning, and test-time learning; and \textit{Multimodal} covers visual fact recall, visual search, visual update, and cross-modal reasoning. Each question is jointly evaluated using LLM-as-a-Judge, F1, and BLEU to reduce biases from any single metric.

\vspace{-0.2cm}
\section{Experiments}
\label{sec:exp}

We evaluate three mainstream memory paradigms. Detailed settings are provided in Appendix~\ref{app:exp_setting}. 

\textbf{Long-Context Agents.} To test whether frontier models can handle long-horizon memory tasks by relying solely on context, these agents concatenate the full interaction history into the prompt as in-context memory, without explicit abstraction, updating, or retrieval. We evaluate GPT-5.4-mini~\cite{openai2026gpt54mini}, Qwen3.5 plus~\cite{qwen35blog}, Gemini 3 flash~\cite{googledeepmind2026gemini3flash}, DeepSeek V4~\cite{deepseekai2026deepseekv4} and Claude Haiku 4.5~\cite{anthropic2025claudehaiku45}. As no independent memory state is exposed, only final question-answering performance is measured.

\textbf{Manually  Designed Memory Systems.} To assess whether explicitly engineered memory mechanisms can improve memory construction, maintenance, retrieval, and downstream use, we evaluate two types of systems. \textit{External memory agents} such as MemGPT~\cite{packer2024memgptllmsoperatingsystems} and Mem0~\cite{mem0} perform information abstraction, consolidation, and retrieval through learned or hand-crafted modules. \textit{Retrieval-augmented generation (RAG) systems} such as UniversalRAG~\cite{yeo2026universalragretrievalaugmentedgenerationcorpora} store historical information in an indexed document store and access it via retrieval. To control for backbone differences, all systems use GPT-5.4-nano~\cite{openai2026gpt54mini} as the base model. Because these systems expose observable memory states and retrieval outputs, the full memory lifecycle can be evaluated.

\textbf{Harness-Based Memory Agents.} To examine whether agents can autonomously manage memory without a fixed external module, we evaluate agent harnesses where memory is written, maintained, retrieved, and used by the harness itself during interaction. We test OpenClaw~\cite{steinberger2025openclaw} paired with GPT-5.4~\cite{gpt54} and DeepSeek-V4, and Codex~\cite{codex2026} paired with GPT-5.4, feeding session contexts sequentially and testing with staged checkpoint QA. Since the internal memory process is difficult to decompose, we primarily conduct end-to-end evaluation.

\begin{table*}[h]
\centering
\setlength{\aboverulesep}{0pt}
\setlength{\belowrulesep}{0pt}
\caption{Performance of baselines on memory quality and question answering (QA) quality.
All values are reported in \% as the mean across samples.
\textbf{Memory metrics:} \textit{Recall} = \textit{Memory Recall}, \textit{Corr} = \textit{Memory Correctness}, \textit{Hallu} = \textit{Memory Hallucination}, \textit{Irrel} = \textit{Memory Irrelevance}, \textit{Update} = \textit{Update Handling}, and \textit{IntRej} = \textit{Interference Rejection}, averaged over samples containing interference items.
\textbf{QA metrics:} \textit{QA-C} = \textit{QA Correct}, \textit{QA-H} = \textit{QA Hallucination}, \textit{QA-O} = \textit{QA Omission}, and \textit{RC} = \textit{Retrieval Coverage}. In the External Memory Agents, the dashed line separates caption-based text only systems above from multimodal systems using both images and text below. All judgments are conducted with GPT-5.4-mini as the evaluator.}
\renewcommand{\arraystretch}{1.38}
\setlength{\tabcolsep}{4pt}
\resizebox{\textwidth}{!}{%
\begin{tabular}{l cccccc cccccc}
\toprule
\rowcolor{HeaderGray}
\textbf{Method}
 & \multicolumn{6}{c}{\textbf{Memory Quality}}
 & \multicolumn{6}{c}{\textbf{QA Quality}} \\
\rowcolor{HeaderGray}
 & \textbf{Recall$\uparrow$} & \textbf{Corr$\uparrow$} & \textbf{Hallu$\downarrow$} & \textbf{Irrel$\downarrow$} & \textbf{Update$\uparrow$} & \textbf{IntRej$\uparrow$} & \textbf{QA-C$\uparrow$} & \textbf{QA-H$\downarrow$} & \textbf{QA-O$\downarrow$} & \textbf{RC$\uparrow$} & \textbf{F1$\uparrow$} & \textbf{BLEU-1$\uparrow$} \\
\midrule
\rowcolor{GroupGreen}
\multicolumn{13}{l}{\textbf{RAG}} \\
Qwen3-VL-Embedding-8B~\cite{zhang2025qwen3} & \underline{86.22} & \textbf{98.15} & \textbf{1.18} & \textbf{0.67} & \textbf{59.02} & 28.21 & 51.86 & 28.02 & \textbf{20.12} & 73.44 & 32.21 & 17.84 \\
UniversalRAG~\cite{yeo2025universalrag} & 84.56 & 96.90 & 2.42 & \underline{0.67} & 57.98 & 27.34 & 39.62 & 31.67 & 28.70 & 60.93 & 27.06 & 14.16 \\
\midrule
\rowcolor{mySoftBlue}
\multicolumn{13}{l}{\textbf{External Memory}} \\
A-Mem~\cite{xu2025mem} & 52.54 & 96.60 & 2.57 & 0.83 & \underline{58.86} & \textbf{58.94} & \underline{54.63} & 22.94 & 22.43 & \underline{74.19} & \textbf{34.40} & \textbf{19.86} \\
MemGPT~\cite{packer2023memgpt} & 85.20 & 96.98 & 2.28 & 0.74 & 58.18 & 25.44 & \textbf{57.81} & \underline{22.05} & \underline{20.14} & \textbf{84.99} & \underline{33.21} & \underline{18.33} \\
SimpleMem~\cite{liu2026simplemem} & 78.84 & 96.96 & 1.44 & 1.35 & 53.43 & 24.79 & 42.93 & 25.60 & 31.47 & 48.03 & 26.00 & 12.30 \\
\cdashline{1-13}
Omni-SimpleMem~\cite{liu2026omnisimplememautoresearchguideddiscoverylifelong} & 58.48 & 72.92 & 15.95 & 9.95 & 52.65 & \underline{43.22} & 43.03 & 32.24 & 24.72 & 62.55 & 25.86 & 12.52 \\
M2A~\cite{feng2026m2amultimodalmemoryagent} & \textbf{86.83} & \underline{97.47} & \underline{1.25} & 1.28 & 56.41 & 23.42 & 50.14 & 29.29 & 20.57 & 64.62 & 31.77 & 17.54 \\
ViLoMem~\cite{bo2026agenticlearnergrowandrefinemultimodal} & 85.96 & 81.61 & 10.65 & 7.74 & 55.73 & 24.93 & 49.77 & 25.20 & 25.02 & 70.71 & 29.51 & 15.63 \\
MIRIX~\cite{wang2025mirixmultiagentmemoryllmbased} & 64.79 & 73.50 & 5.15 & 1.58 & 56.97 & 31.42 & 44.46 & \textbf{20.79} & 34.75 & 61.90 & 24.90 & 12.65 \\
AUGUSTUS~\cite{jain2025augustus} & 84.63 & 96.66 & 2.63 & 0.70 & 57.42 & 28.85 & 42.01 & 32.38 & 25.61 & 57.33 & 27.24 & 13.87 \\
\bottomrule
\multicolumn{13}{l}{\footnotesize \textit{Best in \textbf{bold}, second-best \underline{underlined}.}} \\
\end{tabular}}
\label{tab:data150-gpt-quality}
\end{table*}

\subsection{\textcolor{black}{Main Results}}
Table~\ref{tab:data150-gpt-quality} reports the overall performance of different human designed systems across the full memory lifecycle. We identify four main findings.\ding{182} \textbf{Multimodal memory is still not effectively used.}  Text-based systems such as MemoryGPT and A-Mem achieve more stable final answer quality, while multimodal systems such as ViLoMem and MIRIX show limited downstream gains despite access to visual inputs. This suggests that current systems still struggle to encode and reuse visual evidence as reliable long term memory. \ding{183} \textbf{High memory quality does not necessarily lead to high QA quality.} High memory quality does not necessarily lead to high QA quality. Qwen3-VL-Embedding and M2A perform well in memory storage and recall, but their final answers remain limited. This indicates that correct memory writing is insufficient; systems must also retrieve and use the right evidence during answer generation. \ding{184} \textbf{Retrieval remains a key bottleneck for final performance.} MemoryGPT achieves the strongest evidence retrieval and answer correctness, while A-Mem uses retrieved information effectively despite lower memory coverage. In contrast, AUGUSTUS constructs reasonably good memories but fails to surface key evidence at inference time, limiting its final QA performance. \ding{185} \textbf{Most systems remain weak in memory updating and distractor rejection.}  Nearly all systems are brittle under information changes and interfering content, indicating that they tend to accumulate memories rather than maintain a consistent long-term state. This suggests that current human designed memory systems still focus more on how much they remember than on how well they maintain and update memory over time.

\begin{wraptable}{r}{0.58\textwidth}
\setlength{\aboverulesep}{0pt}
\setlength{\belowrulesep}{0pt}
\centering
\captionsetup{font=small}
\caption{QA quality results for all base models and harness agents. 
\textit{QA-C} denotes \textit{QA Correct}, 
\textit{QA-H} denotes \textit{QA Hallucination}, 
and \textit{QA-O} denotes \textit{QA Omission}.}
\label{tab:data150-gpt-qa-harness}

\scriptsize
\renewcommand{\arraystretch}{0.99 }
\setlength{\tabcolsep}{2.6pt}
\resizebox{\linewidth}{!}{%
\begin{tabular}{l ccccc}
\toprule
\rowcolor{HeaderGray}
\textbf{Method}
 & \multicolumn{5}{c}{\textbf{QA Quality}} \\
\rowcolor{HeaderGray}
 & \textbf{QA-C$\uparrow$} 
 & \textbf{QA-H$\downarrow$} 
 & \textbf{QA-O$\downarrow$} 
 & \textbf{F1$\uparrow$} 
 & \textbf{BLEU-1$\uparrow$} \\
\midrule
\rowcolor{GroupGreen}
\multicolumn{6}{l}{\textbf{Base Model}} \\
Qwen3.5 plus & 51.05 & 16.90 & 32.05 & 21.04 & 8.68 \\
Deepseek V4 & \textbf{69.13} & \textbf{11.46} & \underline{19.41} & \textbf{28.18} & \textbf{13.61} \\
Gemini 3 Flash & 51.69 & 23.69 & 24.62 & \underline{22.93} & 10.32 \\
Claude Haiku 4.5 & 36.71 & 25.47 & 37.83 & 22.05 & \underline{10.79} \\
GPT 5.4-mini & \underline{58.27} & 27.86 & \textbf{13.87} & 21.31 & 8.76 \\
\midrule
\rowcolor{mySoftBlue}
\multicolumn{6}{l}{\textbf{Harness}} \\
Codex-GPT 5.4-nano & \textbf{53.62} & 20.76 & \textbf{25.62} & \textbf{32.56} & 10.12 \\
OpenClaw-DeepSeek V4 & \underline{50.29} & \underline{15.57} & 34.14 & 28.38 & \textbf{18.16} \\
OpenClaw-GPT 5.4-nano & 48.31 & \textbf{19.55} & \underline{32.13} & \underline{30.32} & \underline{15.71} \\
\bottomrule
\end{tabular}}
\label{base}
\end{wraptable}
Table~\ref{base} compares final answer performance between long context agents and harness based memory agents. Most long context agents perform poorly, with some falling below dedicated memory systems, indicating that the benchmark requires long horizon evidence integration rather than context extension alone. DeepSeek V4 benefits mainly from its larger context window, while standard context models remain limited. Harness based memory agents outperform most human designed memory systems, suggesting that agent managed memory is more flexible. However, the same backbone performs differently across harnesses, showing that native memory design and adaptation mechanisms also affect final performance.

\section{\textcolor{black}{Analysis}}

\noindent \textbf{[RQ1] \textit{Where do memory failures occur in the lifecycle?}}

\begin{wrapfigure}{r}{0.43\textwidth} 
  \centering
  \includegraphics[width=\linewidth]{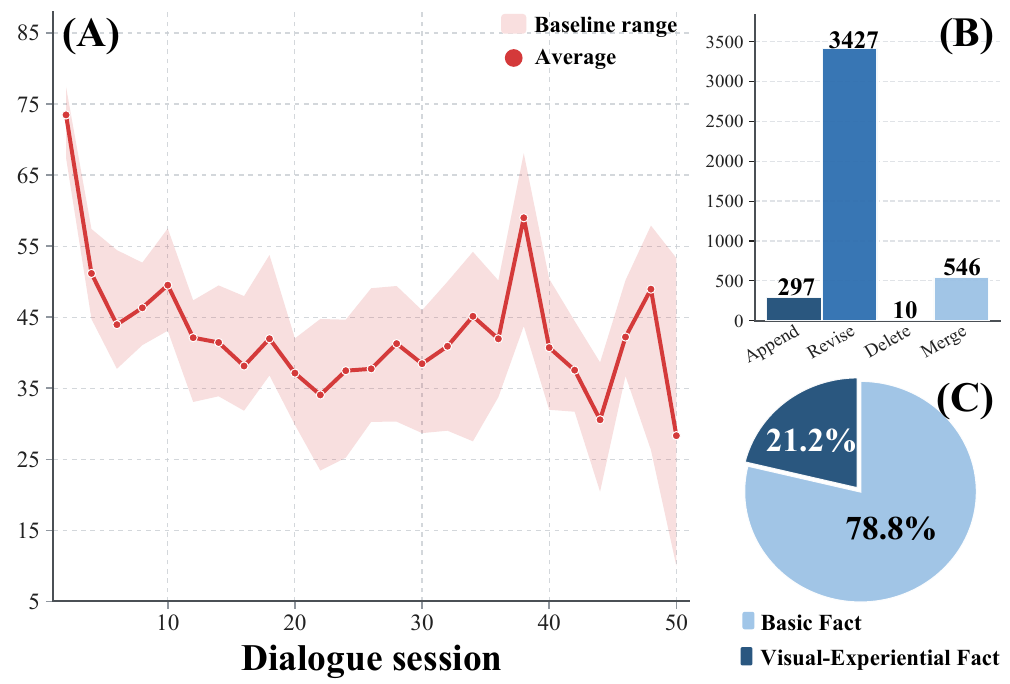}
\caption{
(a) shows the trend of average QA accuracy across dialogue sessions.
(b), (c) summarize memory-point composition in terms of update operations (Append, Revise, Delete, and Merge) and fact salience.
}
\vspace{-3em}
  \label{trade1}
\end{wrapfigure}

\textit{Memory failures occur across the full lifecycle and compound over time.} \textit{(i)} Figure~\ref{trade1}(a) shows that storing more memories does not necessarily make them usable; even with high storage coverage, systems may fail to retrieve the key evidence needed for the current decision. \textit{(ii)} As illustrated in Figure~\ref{trade1} (b), most systems rely on append only updates, adding new information when evidence changes rather than revising, removing, or reorganizing obsolete memories. \textit{(iii)} Over long trajectories, Figure~\ref{trade1}(c) captures a compounding pattern in which early omissions reduce later evidence availability, while incorrect outputs may contaminate future memory updates and further induce hallucinated answers.

\noindent  \textbf{[RQ2] \textit{Are memory system designs constrained by domain-specific data?}}

\begin{figure}[h]
    \centering
        \includegraphics[width=1\linewidth]{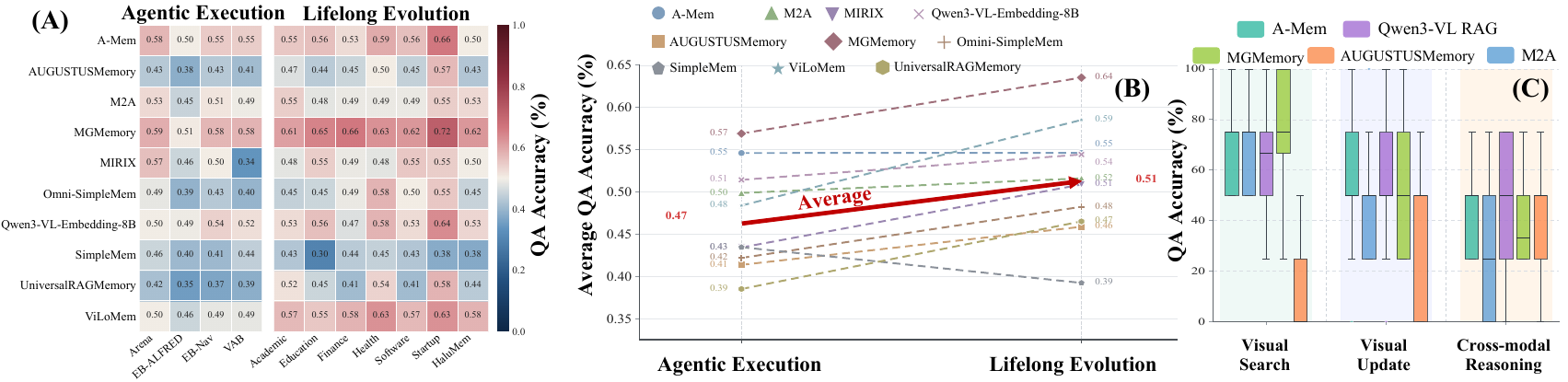}
\caption{\small
Performance comparison across scenarios and visual QA tasks.
(a) Heatmap of baseline performance across fine-grained task categories in the agent-scenario and long-dialogue settings.
(b) Average baseline performance under the two settings, where long-dialogue tasks show higher overall performance than agent-scenario tasks.(c) Box plots of baseline performance across three visual QA task types.
}
    \label{fig:placeholder}
\end{figure}

\textit{Memory performance varies across domains.} As shown in Figure~\ref{fig:placeholder}(a-b), most systems perform better in \textit{Lifelong Evolution} than in \textit{Agentic Execution}. This suggests that existing methods are more suited to explicit long-term state evolution, while extracting usable memory from action traces and environment feedback remains challenging. Performance also differs across tasks, with long-horizon embodied tasks such as visual navigation posing greater challenges, suggesting that current systems still struggle to track memory across sessions and use it for later decisions.

\noindent \textbf{[RQ3] \textit{How does multimodal affect the memory lifecycle?}}

\textit{Memory systems still struggle with complex visual memory tasks.} As shown in Figure~\ref{fig:placeholder}(c), systems perform relatively stably on simple visual fact recall, but degrade on tasks that depend on long interaction histories, such as cross-modal reasoning. This suggests that the core challenge of multimodal memory is to maintain visual states over time and integrate visual evidence with historical context when needed.

\noindent \textbf{[RQ4]  \textit{What strengths and limitations do different memory systems exhibit across task types?}}

\begin{wrapfigure}{r}{0.53\textwidth} 
  \centering
  \includegraphics[width=\linewidth]{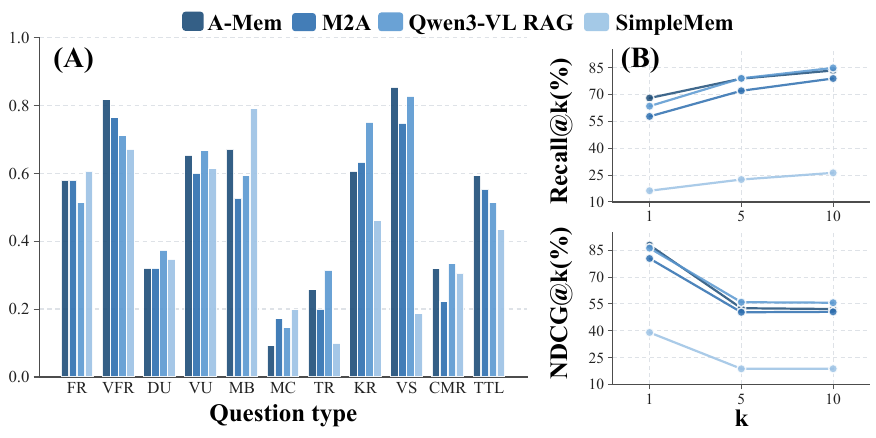}
\caption{
(a) Fine-grained QA performance of different baselines on individual tasks.
(b) Recall@K and NDCG@K (Normalized Discounted Cumulative Gain) trends under different retrieval cutoffs.
}
  \label{backbone}
\end{wrapfigure}

\textit{Memory performance depends more on system design than on backbone scale or retrieval volume} As shown in Table~\ref{tab:data150-gpt-quality}, most systems achieve high memory storage recall and writing quality, yet their evidence recall at question-answering time drops substantially, indicating that correctly stored memories are not effectively surfaced when needed. Figure~\ref{backbone}(b) further shows that increasing the retrieval scope does not always improve answer quality, as longer contexts may introduce redundant, outdated, or irrelevant evidence. This issue is more evident in multimodal tasks, where long interactions create substantial visual redundancy and make key visual evidence harder to locate and use.

\noindent \textbf{[RQ5] \textit{Can agents turn memory into action?}}

\textit{Past memory is not reliably converted into reusable knowledge, and experiential evidence remains fragile.} As shown in \textcolor{black}{Figure~\ref{backbone}(a)}, systems perform worse on reasoning and test-time learning tasks, suggesting that they are better at storing past information than using it to guide future decisions. Analysis of retrieved memory points in Figure~\ref{trade1}(c) shows that retrieved memories are dominated by explicit textual facts, whereas tool feedback, failed actions, and visual details are often omitted.

\noindent \textbf{[RQ6]  \textit{How far do human designed memory systems fall short in agentic memory?}}

\vspace{1cm}
\begin{wrapfigure}{r}{0.45\textwidth} 
  \centering
  \includegraphics[width=\linewidth]{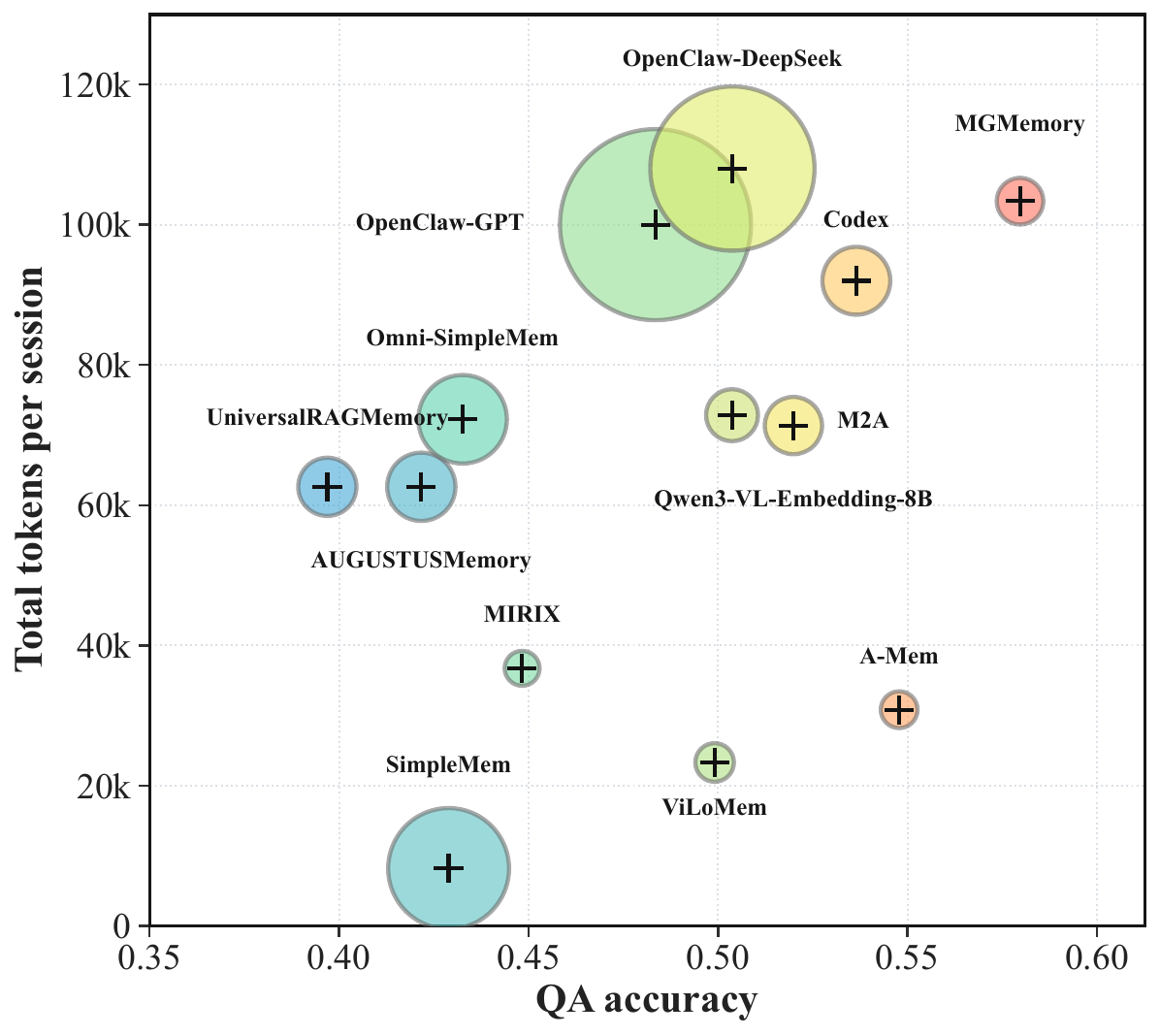}
\caption{\small
Token efficiency \& QA performance trade-off among different baselines.
Circle size indicates average inference time, with larger circles denoting higher time cost.
}
  \label{harness}
  \vspace{-3em}
\end{wrapfigure}

\textit{Fixed memory architectures struggle to adapt to dynamic memory demands.} As shown in Figure~\ref{harness}, human designed memory systems perform comparably to harness based methods on simpler long-horizon tasks, but their fixed pipelines become limiting in complex agentic settings where memory must adapt to task feedback and environmental changes.
Harness based agentic memory managers are more flexible because they can record, retrieve, and revise memory during interaction. However, this result also shows that current harness-based memory remains computationally expensive and framework-dependent, limiting its stability and transferability.

\section{Discussion}
\label{sec:discussion}
The experiments above show that long horizon agent memory remains fragile. Strong storage signals often fail to translate into reliable decisions, and multimodal and interactive settings expose additional failure modes. We distill four directions for future work.

\textbf{\ding{118}  Memory should be shaped through interaction, not fixed as a module.}
Our results show that higher storage quality does not necessarily lead to better performance (Table~\ref{tab:data150-gpt-quality}), while harness-based agents without explicit memory modules outperform some manually designed memory pipelines (Table~\ref{base}). This suggests that effective memory is better understood as a capability shaped by task pressure, rather than as a module that can be optimized in isolation. Future work should explore training paradigms that develop memory through end-to-end interaction objectives.

\textbf{\ding{118}  Memory requires consistent state maintenance, not continuous accumulation.}
Current systems accumulate information but rarely revise or remove obsolete entries (Figure~\ref{trade1}b). Effective memory should be modeled as mutable state that supports revision, conflict resolution, and selective forgetting. New architectures and evaluations are needed that reward state consistency rather than raw coverage.

\textbf{\ding{118}   Effective use of multimodal memory.}
Most systems compress visual observations into textual memory, which often loses spatial, temporal, and procedural details. Our analysis shows that current systems still perform poorly on complex visual tasks, especially when they need to use visual cues and interaction experience for reasoning (Figure~\ref{fig:placeholder}c). Future work should develop architectures that preserve visual memories in usable forms,  with metrics that evaluate whether these memories truly support reasoning and decision-making.

\textbf{\ding{118} Memory evaluation should focus on learning from experience, not retrospective QA.}
Current evaluations often rely on checkpoint QA to measure memory, but the ultimate goal of agent memory is not merely to answer questions about the past, but to improve future behavior. Our experiments show that systems are better at storing facts than at using them for reasoning or learning (Figure~\ref{trade1}a). Future benchmarks should evaluate whether agents can learn from prior experience and failures, rather than merely retrieve past information, and improve behavior across sessions.

\vspace{-0.2cm}
\section{Conclusion}
\label{sec:conclusion}
We presented \textbf{WorldMemArena}, a multimodal multi-session benchmark that evaluates agent memory through the lens of an Action World Interaction Loop. By decomposing memory into four observable stages and annotating each session with gold memory points, updates, and distractors, we enable stage level diagnosis across long context agents, manually designed memory systems, and harness-based memory agents. Experiments show that storage quality alone does not predict final performance, that memory maintenance remains dominated by append only behavior, and that visual evidence is largely reduced to text. These findings suggest that the field should move beyond optimizing memory as a static module and toward developing memory as an adaptive capability grounded in interaction.

\clearpage
\bibliography{main}

\clearpage
\appendix
\section{Experimental Setting}
\label{app:exp_setting}

Unless stated otherwise, every baseline shares the same backbone and
decoding configuration to keep comparisons fair.  The answer-stage and
judge LLMs both run with temperature $0.0$, a maximum completion
budget of $16{,}384$ tokens (which covers reasoning plus output for
GPT-5-class models), and a per-call timeout of $300$\,s, with up to
$10$ concurrent requests.  Backbone variation is controlled at the
model level only: GPT-5.4-mini, Deepseek-V4, Claude Haiku 4.5,
Gemini 3 Flash, and Qwen3.6-plus are evaluated under identical
prompts.  Memory adapters that need an embedding model use OpenAI's
\texttt{text-embedding-3-small} (1{,}536-dim); multimodal retrievers
default to Qwen3-VL-Embedding-8B and the GME Qwen2-VL-2B encoder.
Retrieval is capped at top-$K=10$ items per query for both text and
multimodal paths; the answerer's effective context window is
$128{,}000$ tokens with an $8{,}000$-token reserve for the system and
answer prompt.  Image-augmented QA caps at five images per question
and $45$\,MB of merged payload to stay within provider limits.  The
LLM judge inherits the answer-stage model and runs with
temperature $0.0$ and up to $5$ parallel workers.

\section{Evaluation Metrics}
\label{app:eval-metrics}

\subsection{Notation}
\label{app:eval-notation}

A single evaluation instance corresponds to one trajectory
$\tau = \tau^{(1)} \circ \cdots \circ \tau^{(S)}$ split into $S$
sessions; every per-instance metric below is first aggregated within
$\tau$ and then averaged across instances.  Within session
$\tau^{(s)}$, $\mathcal{D}_s$ collects the \emph{add}/\emph{update}
memory items the policy $\pi$ writes into the memory state $m_t$,
$\mathcal{G}_s$ is the set of gold memory points the system is
expected to remember, and $\mathcal{I}_s$ the set of gold
\emph{interference} points it should reject.  Each gold point
$g \in \mathcal{G}_s$ carries an importance weight $w_g \ge 0$ (default
$1$).  For a QA $q$, $y_q^{\star}$ is the gold answer, $\hat y_q$ the
generated answer, and $\mathcal{V}_q$ the gold evidence points the QA
relies on.  Per-memory and per-QA labels are produced by an LLM judge.

\subsection{Memory metrics}
\label{app:eval-memory}

These metrics decompose ``did the agent build a useful memory'' into
two complementary axes: \emph{coverage} of what should have been
remembered, and \emph{purity} of what was actually stored.  Lifelong
benchmarks also stress two failure modes outside that simple
recall/precision split, namely silently keeping stale facts and
absorbing noise on purpose, so we add \textsc{Update} and
\textsc{IntRej} to capture them.

\begin{itemize}[left=0.2cm]
  \item \textbf{Memory Recall (\textsc{Recall}).}
    Coverage of the gold memory points by the system's add/update
    delta.  An LLM judge decides, semantically, which $g \in
    \mathcal{G}_s$ is supported by some item in $\mathcal{D}_s$;
    let $\mathcal{C}_s \subseteq \mathcal{G}_s$ be the covered subset.
    Recall is importance-weighted because the gold set mixes
    high-stakes facts and incidental details, and we report
    \begin{equation}
      \mathrm{Recall}_s
      = \frac{\sum_{g \in \mathcal{C}_s} w_g}
             {\sum_{g \in \mathcal{G}_s} w_g},
    \end{equation}
    averaged across sessions with $|\mathcal{G}_s|>0$ (sessions with
    no gold are uninformative and dropped).  A semantic judge avoids
    penalising harmless paraphrases or summarisation by the agent.

  \item \textbf{Memory Correctness / Hallucination / Irrelevant
    (\textsc{Corr}, \textsc{Hallu}, \textsc{Irrel}).}
    \textsc{Recall} is blind to garbage: an agent that dumps the
    entire dialogue into memory looks excellent.  We classify each
    stored item $m \in \mathcal{D}_s$ into three exclusive labels:
    \emph{correct} (overall faithful, minor imprecision allowed),
    \emph{hallucination} (partly right but contradicts the dialogue
    on a concrete fact), and \emph{error} (fundamentally wrong, e.g.\
    an event that never happened).  With per-session counts
    $n^{\mathrm{C}}_s, n^{\mathrm{H}}_s, n^{\mathrm{E}}_s$,
    \begin{equation}
      \mathrm{Corr}_s = \frac{n^{\mathrm{C}}_s}{|\mathcal{D}_s|},\quad
      \mathrm{Hallu}_s = \frac{n^{\mathrm{H}}_s}{|\mathcal{D}_s|},\quad
      \mathrm{Irrel}_s = \frac{n^{\mathrm{E}}_s}{|\mathcal{D}_s|},
    \end{equation}
    averaged across sessions with $|\mathcal{D}_s|>0$.
    \textsc{Hallu} and \textsc{Irrel} are reported as
    ``lower is better'': they expose the price an agent pays for a
    high \textsc{Recall}.

  \item \textbf{Update Handling (\textsc{Update}).}
    Long-horizon memory must overwrite stale facts when the world
    changes (e.g., the user moves house).  For every gold update we
    inspect the post-session memory snapshot and label it as
    \emph{updated} (only the new fact is kept), \emph{both} (new and
    old coexist), or \emph{outdated} (only the old fact survives).
    Pooling counts across the sessions of an instance,
    \begin{equation}
      \mathrm{Update}
      = \frac{1.0\cdot N_{\mathrm{updated}}
            + 0.5\cdot N_{\mathrm{both}}
            + 0.0\cdot N_{\mathrm{outdated}}}
             {N_{\mathrm{total}}}.
    \end{equation}
    The half-credit on \emph{both} reflects that the agent has the new
    fact but failed to invalidate the old one; downstream QA can still
    surface the wrong answer.

  \item \textbf{Interference Rejection (\textsc{IntRej}).}
    Real conversations contain casual remarks, jokes, and corrections
    that the agent should \emph{not} commit to memory.  For every gold
    interference point $g \in \mathcal{I}_s$ the post-session snapshot
    is classified as \emph{rejected} or \emph{memorized}, and
    \begin{equation}
      \mathrm{IntRej}
      = \frac{N_{\mathrm{rejected}}}
             {N_{\mathrm{rejected}} + N_{\mathrm{memorized}}}.
    \end{equation}
    A high \textsc{Recall} paired with low \textsc{IntRej} is the
    signature of an indiscriminate writer that hoards everything; the
    two metrics together separate selective memory from a transcript.
\end{itemize}

\subsection{QA metrics}
\label{app:eval-qa}

The memory metrics above audit the memory store directly.  The QA
metrics measure the downstream effect: given the memory the agent
built, can it answer questions whose evidence is no longer in the
local context?  For every QA $q$, the judge compares $\hat y_q$
against $y_q^{\star}$ and the gold evidence list $\mathcal{V}_q$ and
emits a single label
$\ell_q \in \{\mathrm{Correct}, \mathrm{Hallucination},
\mathrm{Omission}\}$; let $n^\star$ be the number of QAs in the
instance that received a valid label.

\begin{itemize}[left=0.2cm]
  \item \textbf{QA Correct / Hallucination / Omission
    (\textsc{QA-C}, \textsc{QA-H}, \textsc{QA-O}).}
    The three labels separate the qualitatively different ways an
    answer can fail: confident-but-wrong (\emph{Hallucination}) is
    treated separately from refusal or ``I don't know''
    (\emph{Omission}), since they imply different failure modes of the
    memory pipeline.
    \begin{align}
      \mathrm{QA\text{-}C} &= \frac{|\{q : \ell_q = \text{Correct}\}|}{n^\star}, \\
      \mathrm{QA\text{-}H} &= \frac{|\{q : \ell_q = \text{Hallucination}\}|}{n^\star}, \\
      \mathrm{QA\text{-}O} &= \frac{|\{q : \ell_q = \text{Omission}\}|}{n^\star}.
    \end{align}

  \item \textbf{Answer F$_1$.}
    The judge label is binary at the QA level; F$_1$ adds a
    fine-grained surface-form signal that captures partial overlap on
    short factual answers.  We tokenise both answers with a normaliser
    that lowercases, drops the stopwords \texttt{a}/\texttt{an}/\texttt{the}/\texttt{and},
    strips punctuation while preserving decimals, and applies Porter
    stemming.  Writing $\widetilde{T}(\cdot)$ for the resulting token
    multiset and $C_q = \widetilde{T}(\hat y_q) \cap \widetilde{T}(y_q^\star)$,
    \begin{equation}
      P_q = \frac{|C_q|}{|\widetilde{T}(\hat y_q)|},\quad
      R_q = \frac{|C_q|}{|\widetilde{T}(y_q^\star)|},\quad
      F_{1,q} = \frac{2 P_q R_q}{P_q + R_q}\quad(0\text{ if }|C_q|=0).
    \end{equation}
    Stemming reduces the penalty for harmless inflection
    (``walk''/``walked'') and is appropriate at the answer-string
    level.

  \item \textbf{BLEU-1.}
    BLEU-1 (unigram BLEU with add-$\epsilon$ smoothing) is reported
    alongside F$_1$ as a precision-leaning surface metric: it weights
    repeated terms and is less generous to padding, so the gap between
    F$_1$ and BLEU-1 is informative on its own.  Tokenisation uses the
    same normaliser \emph{without} Porter stemming, so BLEU-1 stays
    comparable to standard implementations.
\end{itemize}

\subsection{Retrieval metrics}
\label{app:eval-retrieval}

Memory and QA quality measure ``what was stored'' and ``what was
answered''; the retrieval metrics measure the bridge between them,
i.e.\ whether the relevant past evidence is actually surfaced when a
question is asked.  For every QA $q$ the system returns an ordered
list of retrieved items $\mathbf r_q = (r_{q,1}, r_{q,2}, \ldots)$
against the gold evidence set $\mathcal{V}_q$.  We use a soft match
predicate $\mathbf{1}\{r \approx g\}$ that returns $1$ when (i) the
gold memory id is contained in the retrieved item's identifiers, (ii)
the source-session id parsed from the gold matches the session that
contributed $r$, or (iii) the normalised gold content is a substring
of, or has $\ge 0.75$ token-overlap ratio with, the normalised
retrieved text.  These three rules absorb superficial id mismatches
between heterogeneous baselines and avoid awarding credit purely on
verbatim string equality.  Let $\mathrm{cov}_K(q) = \{g \in
\mathcal{V}_q : \exists\, k\le K,\; r_{q,k} \approx g\}$.

\begin{itemize}[left=0.2cm]
  \item \textbf{Retrieval Coverage (\textsc{RC}).}
    A rank-agnostic, semantic-level check: an LLM judge reads the
    full top-$K$ list and decides how many gold evidence points are
    supported anywhere in it.  Letting $|Q|$ be the QA count of the
    instance,
    \begin{equation}
      \mathrm{RC}
      = \frac{1}{|Q|} \sum_{q \in Q}
        \frac{\mathrm{covered}_q}{|\mathcal{V}_q|},
    \end{equation}
    where $\mathrm{covered}_q$ is the judge's count.  \textsc{RC}
    captures retrieval quality without committing to a particular
    rank position, since an answer can succeed as long as the evidence
    is present and the answerer reads the list.

  \item \textbf{Recall@$K$.}
    A strict, rank-bounded counterpart of \textsc{RC} based on the
    soft match predicate (no judge, deterministic).  It probes whether
    the top of the list alone is informative:
    \begin{equation}
      \mathrm{Recall@}K_q = \frac{|\mathrm{cov}_K(q)|}{|\mathcal{V}_q|},
      \qquad K \in \{1, 5, 10\}.
    \end{equation}
    The $K=1$ value is the harshest: it rewards retrievers that put
    the right evidence \emph{first} rather than somewhere in the top
    decile.

  \item \textbf{NDCG@$K$.}
    Recall@$K$ ignores ranking inside the top-$K$.  NDCG@$K$ closes
    that gap by discounting later ranks.  We first turn the retrieval
    list into a binary relevance vector $\boldsymbol\rho_q$ by greedy
    assignment, so that one retrieved item cannot earn credit for two
    golds:
    \begin{equation}
      \rho_{q,k} =
      \begin{cases}
        1, & r_{q,k} \text{ matches a gold not yet covered by ranks } 1{:}k{-}1, \\
        0, & \text{otherwise}.
      \end{cases}
    \end{equation}
    The DCG aggregates this vector with a logarithmic rank discount,
    \begin{equation}
      \mathrm{DCG}_K(\boldsymbol\rho_q)
      = \rho_{q,1} + \sum_{k=2}^{K} \frac{\rho_{q,k}}{\log_2(k+1)}.
    \end{equation}
    The ideal DCG corresponds to all golds appearing as early as
    possible.  Writing $K^{*} = \min(|\mathcal{V}_q|, K)$ for the
    number of golds reachable in the top-$K$,
    \begin{equation}
      \mathrm{IDCG}_K
      = 1 + \sum_{k=2}^{K^{*}} \frac{1}{\log_2(k+1)}.
    \end{equation}
    NDCG@$K$ is the ratio of the two, with the convention that QAs
    with no gold contribute $0$:
    \begin{equation}
      \mathrm{NDCG@}K_q
      = \frac{\mathrm{DCG}_K(\boldsymbol\rho_q)}{\mathrm{IDCG}_K(\boldsymbol\rho_q)}
      \quad (0 \text{ if } |\mathcal{V}_q|=0).
    \end{equation}
\end{itemize}

\subsection{Per-question-type accuracy}
\label{app:eval-qtype}

Aggregate accuracy hides systematic strengths and weaknesses, so we
also report \textsc{QA-C} restricted to QAs of a single semantic type
$t$.  Each gold QA is annotated with one of eleven mutually exclusive
types, grouped along four skill axes summarised in
Table~\ref{tab:qa-types}.

\begin{table}[t]
\centering
\caption{The eleven semantic axes used for per-type QA accuracy.
Each axis is a mutually exclusive label assigned to every gold QA.}
\label{tab:qa-types}
\renewcommand{\arraystretch}{1.20}
\setlength{\tabcolsep}{6pt}
\resizebox{\textwidth}{!}{%
\begin{tabular}{l l l p{0.55\linewidth}}
\toprule
\textbf{Group} & \textbf{Abbr.} & \textbf{Type} & \textbf{What the question tests} \\
\midrule
Basic       & \texttt{FR}  & Fact Recall          & Retrieve a single concrete fact stated earlier in the trajectory. \\
\midrule
Robustness  & \texttt{DU}  & Dynamic Update       & The queried fact has been overwritten later; the answer must reflect the latest version. \\
            & \texttt{MB}  & Memory Boundary      & The answer is not present in memory; the system must abstain rather than fabricate. \\
            & \texttt{MC}  & Memory Conflict      & Two memory items disagree; the system must resolve the conflict using context. \\
\midrule
Reasoning   & \texttt{TR}  & Temporal Reasoning   & The answer requires reasoning about timing, ordering, or duration of events. \\
            & \texttt{KR}  & Knowledge Reasoning  & The answer combines stored facts with general world knowledge. \\
            & \texttt{TTL} & Test-Time Learning   & The system must apply a rule or skill it was taught earlier in the trajectory. \\
\midrule
Multimodal  & \texttt{VFR} & Visual Fact Recall   & The gold fact is anchored to a specific image in memory. \\
            & \texttt{VS}  & Visual Search        & The answer requires locating an object or attribute across visual memory. \\
            & \texttt{VU}  & Visual Update        & A previously observed visual state has changed later in the trajectory; the answer must reflect the most recent observation. \\
            & \texttt{CMR} & Cross-modal Reasoning & The answer combines textual and visual memory. \\
\bottomrule
\end{tabular}}
\end{table}

For each axis $t$, the cell value is \textsc{QA-C} computed only over
QAs of that type, averaged across instances that contain at least one
QA of type $t$.  The \emph{Avg.} column is the unweighted mean of the
eleven per-type values per instance, which prevents types with more
QAs from dominating the headline number.

\section{Additional dataset details}
\label{app:data}

\subsection{Data sources}

Our trajectories are sourced from four upstream agent benchmarks,
including EmbodiedBench~\cite{yang2025embodiedbenchcomprehensivebenchmarkingmultimodal},
VisualAgentBench~\cite{liu2024visualagentbenchlargemultimodalmodels}, the Agent-Arena GUI task
collection~\cite{kadi2025agentarenageneralframeworkevaluating}, together
with an in-house long-horizon dialogue collection that we release alongside this benchmark.

\subsection{Quality validation}
Each generated session passes through automatic validators (memory point
coverage, image caption coverage, interference detectability, update
chain consistency) before being assembled into the dataset. Samples
failing any validator are regenerated up to 3 times.

\subsection{Further Introduction to Dataset Domains}

\paragraph{Lifelong evolution.}
\textbf{Lifelong Evolution} instantiates the lifelong dimension of WorldMemArena  through \textbf{two complementary domains}, specified in the next two paragraphs.
In both domains, experience arrives as an \emph{ordered} sequence of sessions (for example \texttt{S00}, \texttt{S01}, \ldots), and each stage may introduce new observations that \emph{supersede} facts that previously held.
Fine grained supervision comes from staged memory point annotations (including update flags, importance, and, when applicable, superseded ``original'' memories).
From these we derive a cumulative \emph{gold} memory state per session for analysis and scoring.
Evaluation is interleaved through \texttt{qa\_checkpoints} tied to \texttt{covered\_sessions}.
The model is examined only after a stretch of new experience, rather than by replaying the full chat log in a single prompt.
The design targets \textbf{evolving personal state} (identity, relationships, and preferences revealed in \texttt{S00} and later turns) and \textbf{evolving task state} (work outcomes, projects, constraints, and domain milestones) under temporal noise and interference.
This is not static persona QA on a single conversation.

\paragraph{Professional verticals domain.}
The first lifelong domain is organized into six professional verticals (for example academic, software, health, finance, education, startup), with 18 samples in total.
Each trajectory foregrounds a long arc centered on \emph{tasks} (research programs, product delivery, clinical or business workflows).
Professional artifacts and constraints shift over time.
Checkpoint questions may anchor evidence in \textbf{multimodal} references.
Besides memory point identifiers, gold references may include \textbf{image} identifiers tied to per turn attachments, corresponding to documents, interfaces, or scene captures that accompany narrated actions.

\paragraph{Holistic life course domain.}
The second lifelong domain adopts a holistic life course setting with 20 trajectories.
Each trajectory explicitly separates main arc sessions (career and life goal progression) from side arc sessions (daily life, family, health), with per session labels for arc role, event type, and whether the session lies on the primary storyline.
Gold QA evidence in this domain is recorded primarily as text memory point identifiers, emphasizing narrative memory under rich personal context rather than professional domains stratified by category. The two lifelong domains share the same data shape oriented toward \emph{evaluation} (ordered sessions, staged memory points, checkpoint QA), so one lifelong runner and gold state machinery apply throughout Lifelong Evolution.

\paragraph{Agent domain.}
The Agent domain covers long horizon \emph{agent trajectories} in WorldMemArena.
At each step the evaluated model receives an observation, internal reasoning, an executed action, environment feedback, and optional screenshots from diverse simulated or instrumented settings (for example navigation, embodied manipulation, and desktop GUI tasks).
Here the Action World is explicit in the record.
State changes are governed by actions and feedback, not by conversational stance alone, and staged memory point annotations track evolving quantities such as inventory, location, task phase, and failure or success signals.
Probes and post hoc questions therefore target whether memory captures how the environment changed across steps, including updates and interference, rather than surface repetition of phrasing.
In short, the Agent domain instantiates the Action World Interaction Loop in its most direct form.
The trajectory is already a time ordered log of acting upon a world and reading consequences back.

\paragraph{Action World Interaction Loop versus pure long dialogue memory.}
We unify the two lifelong domains and the Agent domain under an Action World Interaction Loop.
In the professional verticals and life course domains, dialogue between the user and the assistant is the \emph{surface channel}.
Each session is anchored to \emph{events in a world} (career moves, compliance deadlines, household logistics, health episodes, material outcomes) that change what is true thereafter, with per turn attachments as observable traces of those events (forms, screenshots, records).
In the Agent domain, the same logic appears without mediation through narration of a human life in natural language.
Observations and screenshots are already traces of an acting agent coupled to an environment.
All three domains require integrating symbolic state evolution with visual grounding where images appear, rather than only summarizing conversational tone or entity mentions.
By contrast, classical long dialogue benchmarks largely test recall cued by \emph{lexical overlap} in extended chat.
They seldom commit to a jointly evolving external task state that can be superseded, or to staged interference and multimodal evidence aligned with what actually happened outside the text channel.
Under this loop, success requires maintaining a \emph{latent world model} of consequences and updates across time.
The evaluated model must remember not only \emph{what was said}, but also \emph{what became true} after actions and outcomes accumulate in a persistent situation.

\section{Adapter interface}
\label{app:adapter}
Every memory system implements the seven-method \texttt{MemoryAdapter}
interface: \texttt{reset}, \texttt{ingest\_turn}, \texttt{end\_session},
\texttt{snapshot\_memories}, \texttt{export\_memory\_delta},
\texttt{retrieve}, \texttt{get\_capabilities}. This unifies systems
written in Python, hosted via local servers (Qdrant, Neo4j), or wrapped
from external repositories.

\section{More Experiment}

\begin{figure}
    \centering
    \includegraphics[width=0.5\linewidth]{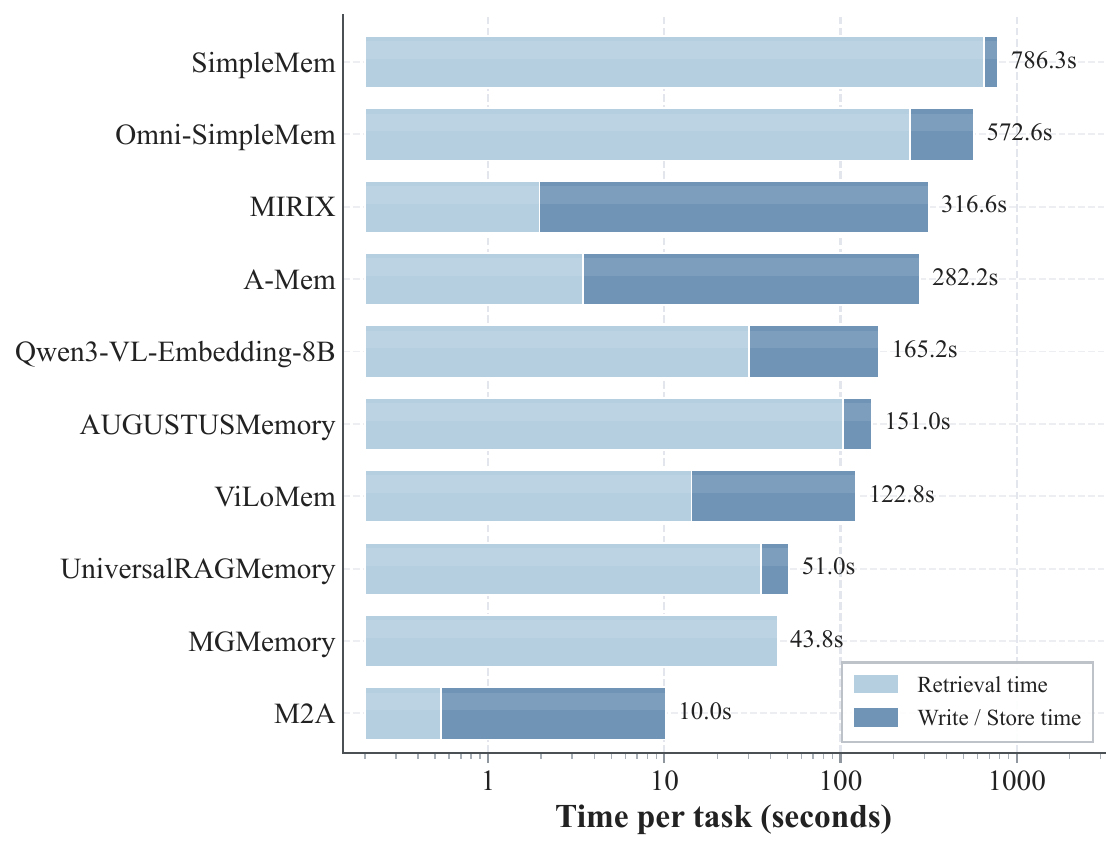}
\caption{%
Mean per-task wall-clock time on a log-scale axis, split into
retrieval (light blue) and write/store (dark blue).
}
\label{fig:latency}
\end{figure}

\paragraph{Latency profile of memory baselines.}
Figure~\ref{fig:latency} reports the mean per-task wall-clock time of
each memory method, split into retrieval and write/store phases.
Total cost spans almost two orders of magnitude, from M2A
($10.0$\,s) to SimpleMem ($786.3$\,s), and the split between the two
phases differs substantially across designs.  Read-heavy methods
such as SimpleMem and Omni-SimpleMem spend the bulk of their budget
re-scanning the dialogue at query time, whereas write-heavy methods
such as MIRIX and A-Mem front-load the cost during ingestion and
then serve queries in milliseconds; MGMemory pushes this pattern to
its limit by indexing inline, so its write phase is effectively free
($\approx 2$\,ms).  Write and retrieval time are therefore largely
independent design choices, and the cost frontier is occupied by
methods that keep \emph{both} small (M2A, MGMemory).  In other
words, latency on long-horizon traces is dominated by the memory
strategy rather than by raw backbone speed: choosing where to pay,
ingestion or query, has a far larger impact than choosing the LLM.

\paragraph{Retrieval on WorldMemArena.}
Table~\ref{tab:data150-gpt-retrieval} separates methods by paradigm.
Dense RAG with Qwen3-VL-Embedding-8B achieves the strongest ranking quality at larger $K$, with the highest NDCG@5 / NDCG@10, indicating well-ordered candidate lists beyond the first hit.
UniversalRAG lags on both recall-oriented metrics and NDCG, suggesting weaker coverage or ranking on this split.
Among agent-memory systems, MemoryGPT reaches very high Recall@K and NDCG@1, on par with the Base Model rows where $\mathrm{R@1}=\mathrm{R@5}=\mathrm{R@10}$.
That saturation pattern is consistent with frequent early retrieval of the relevant unit, while graded relevance within the top-$K$ list remains difficult: MemoryGPT does not surpass the embedding baseline on NDCG@5 / NDCG@10 despite extreme recall.
A-Mem occupies a different regime, with moderately high recall and second-best NDCG@5 / NDCG@10 among memory methods, which highlights a trade-off between hit rate and graded relevance across memory designs.

\paragraph{Variance across memory architectures.}
Beyond the top rows, the agent-memory block exhibits large spread.
SimpleMem and MIRIX are substantially weaker, indicating that lightweight or misaligned memory indexing fails this retrieval benchmark.
Omni-SimpleMem and M2A recover much of the gap toward mid-tier recall and NDCG, while ViLoMem remains weaker at small $K$ despite improving at R@10.
AUGUSTUS tracks A-Mem on recall but does not translate into superior NDCG@$K$, reinforcing that high recall alone is insufficient when evaluation stresses ranking quality.

\paragraph{Backbone competence along 11 types of capabilities.}
Table~\ref{tab:diff-sys-exp-backbones} clarifies how backbone choice can interact with memory/RAG behavior.
Deepseek~V4 attains the highest average and leads Fact Recall, Memory Boundary, Memory Conflict, and most multimodal axes (Visual Fact Recall, Visual Search, Visual Update), suggesting stronger grounding and visual evidence use under the benchmark definitions.
GPT~5.4-mini is second on average with peaks on Temporal Reasoning, Knowledge Reasoning, Test-Time Learning, and Cross-modal Reasoning, but it collapses on Memory Boundary, indicating reasoning-centric strength paired with weak explicit boundary control.
Claude Haiku~4.5 excels on Dynamic Update and Test-Time Learning while suffering on multimodal retrieval scores (Visual Search in particular).
Gemini~3 Flash and Qwen3.5 Plus are more balanced but below the top two on average, with Gemini especially weak on Temporal Reasoning and Visual Search.
Together, the two tables support a systems-level reading: leaderboard differences at retrieval time reflect both pipeline design and the backbone's axis-wise strengths, especially when multimodal alignment or boundary-sensitive memory behavior is required.

\section{More Analysis}
Due to space limitations, the main text cannot provide a detailed analysis. Here, we provide an extended analysis following RQ1–RQ6 in the main paper.

\paragraph{\ding{118}~ Long-horizon collapse.} 
\textit{\textbf{Lifecycle failures compound into long-horizon memory collapse.}} Lifecycle failures compound as trajectories become longer. Early omissions in writing reduce the evidence available to later retrieval. Retrieval failures then prevent the model from grounding later answers, and incorrect answers may further pollute subsequent memory updates. This creates a snowball effect in which later-session reasoning questions become increasingly difficult, even when the required evidence was present earlier in the trajectory. The degradation is particularly severe for reconstructed agentic worlds, where the system must remember not only explicit statements, but also actions, tool outcomes, visual states, and causal consequences. 

\textbf{\ding{118}~  Agentic trajectories expose domain brittleness.} Reconstructed agentic worlds remain challenging for most systems because therelevant evidence is distributed across dense action sequences, tool feedback,GUI states, screenshots, and environment transitions. Many existing systemsrely on assumptions that work well for static conversations or document-likehistories, but break down when memory must be extracted from interactiveexperience. In agentic trajectories, the system must decide which actionsmattered, which failures should be remembered, which object states changed,and which tool outcomes should guide future behavior. As a result, systemsthat perform well on prior memory benchmarks may show a sharp forgetting curveunder more interactive and causally dense settings.

\textbf{\ding{118}~   Retrieval is limited by precision and text bias.}
Human-designed memory systems show a clear trade-off between recall and
precision. Increasing the retrieval budget can improve the chance of including
relevant evidence, but it does not necessarily improve final answers. Larger
retrieved contexts may introduce outdated, conflicting, or irrelevant
memories, making it harder for the model to identify the correct evidence.
This indicates that retrieval quality cannot be reduced to retrieving more
items; effective memory systems require query-aware selection, evidence
ranking, and conflict filtering. This limitation is even more pronounced in
multimodal tasks. Many systems store images or screenshots at a surface level,
but retrieval still relies heavily on text proxies such as captions, OCR, or
generated summaries. As a result, visual evidence is only usable if it was
correctly textualized during writing. Current multimodal memory therefore
remains largely text-centric, highlighting the need to preserve visual
evidence as first-class information rather than reducing it to incomplete
textual descriptions.

\textbf{\ding{118}~  Past experience is not automatically reusable.}
A key goal of agent memory is not only to answer questions about the past,
but also to improve future behavior. Our results suggest that this ability
remains limited. Systems can often repeat explicit facts from earlier
sessions, but they struggle to convert past experiences into action-guiding
knowledge. This is most evident in reasoning and test-time learning tasks,
where the system must infer a reusable rule, remember a previous failure, or
adapt its future decision based on earlier feedback. In other words, current
memory systems are better at recalling past information than at turning that
information into future decisions.

\textbf{\ding{118}~  Experiential evidence is fragile.}
Qualitative cases show that tool feedback, failed actions, visual details,
and implicit causal lessons are among the easiest information types to lose.
In contrast, explicit textual facts are much easier to write and retrieve.
This asymmetry creates a gap between factual memory and agentic memory: a
system may remember what a user said, while failing to remember what happened
when it acted, why an attempt failed, or which strategy succeeded. The same
issue also appears in long dialogue, where systems often preserve local facts
but fail to consolidate them into higher-level user models or stable cognitive
states. These findings suggest that action-oriented memory requires more than
storage and retrieval; it requires transforming experience into reusable
policies, constraints, feedback patterns, and decision priors.

\textbf{\ding{118}~ Limits of human-designed memory systems.}
Human-designed memory systems provide useful structure, but they also impose
fixed assumptions about what should be stored, how memory should be organized,
and how retrieval should operate. These assumptions can work well in narrow
settings, yet become limiting in agentic environments where useful memory
depends on the task, tool feedback, visual state, and future action needs.
The main weakness is not only lower absolute performance, but also reduced
adaptability: a memory pipeline tuned for one domain may not know how to
reorganize itself when the environment changes.

\end{document}